\pdfoutput=1

\documentclass[11pt]{article}

\usepackage[final]{acl}

\usepackage{times}
\usepackage{latexsym}

\usepackage[T1]{fontenc}

\usepackage[utf8]{inputenc}

\usepackage{microtype}

\usepackage{inconsolata}

\usepackage{graphicx}

\usepackage{booktabs}
\usepackage{multirow}
\usepackage{graphicx}

\usepackage{algorithm}
\usepackage{algorithmic}

\usepackage{newfloat}
\usepackage{listings}

\usepackage{subfigure}
\usepackage{amsthm,amsmath,amssymb} 
\usepackage{mathrsfs}
\usepackage{lineno}
\usepackage{xcolor}
\usepackage{color, soul}
\usepackage[inline]{enumitem}
\usepackage{acronym}
\usepackage{enumitem}

\usepackage[skip=5pt]{caption}
\acrodef{LLM}{large language model}
\acrodef{RLAIF}{reinforcement learning from AI feedback}
\acrodef{RLHF}{reinforcement learning from human feedback}
\acrodef{DPO}{direct preference optimization}
\acrodef{PRO}{preference ranking optimization}

\acrodef{Fine-grained PA}{Fine-grained Preference Alignment}
\acrodef{NLP}{natural language processing}
\acrodef{KnowTuning}{knowledge-aware fine-tuning}
\acrodef{SFT}{supervised fine-tuning}
\acrodef{SPO}{Subject-Predicate-Object}
\acrodef{QA}{question answering}
\acrodef{PPO}{Proximal Policy Optimization}
\acrodef{KILT}{knowledge intensive language tasks}

\newcommand{\header}[1]{\vspace{1.5mm}\noindent\textbf{#1}.}

\title{KnowTuning: Knowledge-aware Fine-tuning for Large Language Models}

\author{Yougang Lyu\textsuperscript{\rm 1,3} \qquad Lingyong Yan\textsuperscript{\rm 2} \qquad Shuaiqiang Wang\textsuperscript{\rm 2}\qquad  Haibo Shi\textsuperscript{\rm 2}\qquad  Dawei Yin\textsuperscript{\rm2}  \\{\bf Pengjie Ren\textsuperscript{\rm 1}\qquad Zhumin Chen\textsuperscript{\rm 1}\qquad Maarten de Rijke\textsuperscript{\rm 3}\qquad} {\bf Zhaochun Ren}\textsuperscript{\rm 4}\thanks{\ Corresponding author.} 
\\
        \textsuperscript{\rm 1}Shandong University, Qingdao, China \qquad \textsuperscript{\rm 2}Baidu Inc., Beijing, China
\\ \textsuperscript{\rm 3}University of Amsterdam, Amsterdam, The Netherlands 
\\ \textsuperscript{\rm 4}Leiden University, Leiden, The Netherlands
\\
youganglyu@gmail.com, \{yanlingyong, wangshuaiqiang\}@baidu.com
\\
haiboshi@outlook.com, yindawei@acm.org, jay.ren@outlook.com\\
chenzhumin@sdu.edu.cn, m.derijke@uva.nl, z.ren@liacs.leidenuniv.nl
}

\begin{document}

\maketitle


\begin{abstract}
Despite their success at many \ac{NLP} tasks, \acp{LLM} still struggle to effectively leverage knowledge for knowledge-intensive tasks, manifesting limitations such as generating incomplete, non-factual, or illogical answers. 
These limitations stem from inadequate knowledge awareness of \acp{LLM} during vanilla fine-tuning.
To address these problems, we propose a \ac{KnowTuning} method to improve fine-grained and coarse-grained knowledge awareness of \acp{LLM}. 
We devise a fine-grained knowledge augmentation stage to train \acp{LLM} to identify difficult fine-grained knowledge in answers. We also propose a coarse-grained knowledge comparison stage to train \acp{LLM} to distinguish between reliable and unreliable knowledge, in three aspects: completeness, factuality, and logicality.
Extensive experiments on both generic and medical \ac{QA} datasets confirm the effectiveness of \ac{KnowTuning}, through automatic and human evaluations, across various sizes of \acp{LLM}. We further verify
that KnowTuning generates more facts with less factual error rate under fine-grained facts evaluation.
\end{abstract}


\section{Introduction}
\Acfp{LLM} have become a default solution for many \acf{NLP} scenarios, including the \acf{QA} task~\cite{DBLP:conf/nips/BrownMRSKDNSSAA20,DBLP:conf/nips/Ouyang0JAWMZASR22,DBLP:conf/emnlp/QinZ0CYY23}. 
To achieve strong performance, most \ac{LLM} first accumulate substantial knowledge by pre-training on extensive datasets~\cite{DBLP:journals/corr/abs-2310-06825,DBLP:journals/corr/abs-2307-09288}.
Then, in the \acf{SFT} stage, these \acp{LLM} further learn downstream domain knowledge and how to exploit the corresponding knowledge to answer diverse questions~\cite{DBLP:conf/iclr/WeiBZGYLDDL22,DBLP:journals/corr/abs-2210-11416,DBLP:conf/acl/WangKMLSKH23,DBLP:journals/corr/abs-2304-03277,DBLP:journals/corr/abs-2305-18395,DBLP:journals/corr/abs-2308-13259}.
\begin{figure}[t]
  \centering
    \subfigure[Fine-grained knowledge awareness.]{
\centering
\includegraphics[width=1.0\linewidth]{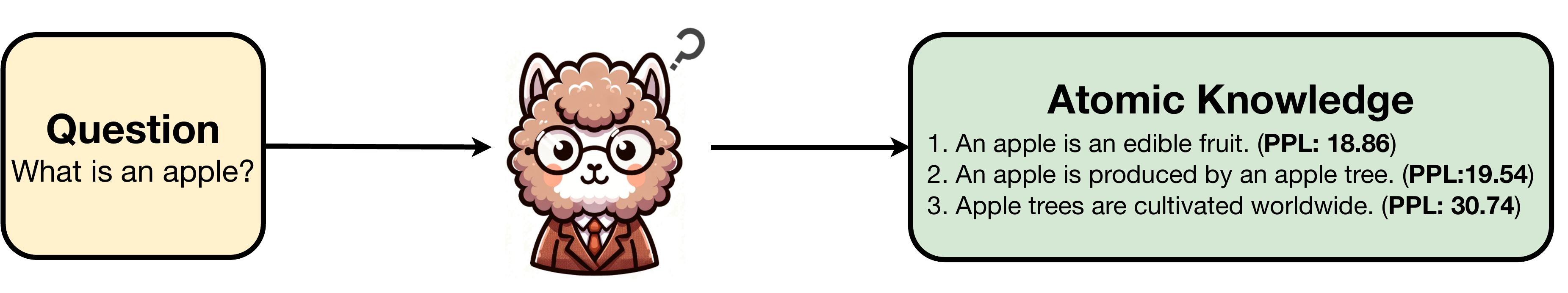}
\label{fig:3a}
}%

\vspace*{-2mm}

\subfigure[Coarse-grained knowledge awareness.]{
\centering
\includegraphics[width=1.0\linewidth]{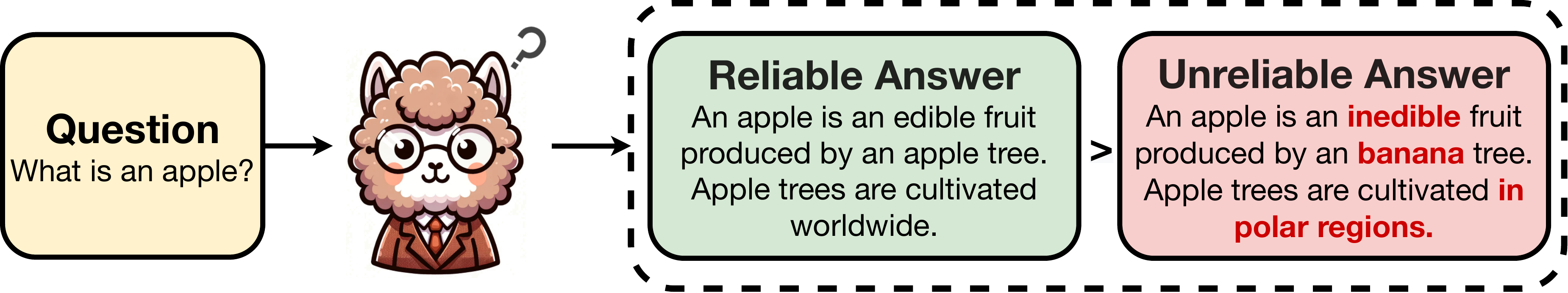}
\label{fig:3b}
}
\vspace*{-2mm}  
\caption{Illustrations of vanilla fine-tuned \acp{LLM} lacking knowledge awareness. (a) Vanilla fine-tuned \acp{LLM}  struggles to identify the fine-grained knowledge to answer a specific question precisely. (b) Vanilla fine-tuned \acp{LLM} cannot effectively distinguish between reliable knowledge and unreliable knowledge in answers.}
\label{fig:example}
\vspace*{-6mm}  
\end{figure}

However, fine-tuned LLMs often struggle to effectively leverage knowledge for complex knowledge-intensive question-answering~\cite{DBLP:journals/corr/abs-2306-09296,DBLP:journals/corr/abs-2310-09725,DBLP:journals/corr/abs-2310-00741,DBLP:journals/corr/abs-2307-03109}. Concretely, many recent studies indicate that \acp{LLM} are susceptible to generating incomplete answers, offering incomprehensive and insufficient knowledge~\cite{DBLP:journals/corr/abs-2212-13138,DBLP:journals/corr/abs-2303-16421,DBLP:conf/acl/XuSIC23}; non-factual answers, delivering factually incorrect knowledge~\cite{DBLP:journals/corr/abs-2305-12421,DBLP:conf/emnlp/MinKLLYKIZH23,DBLP:journals/corr/abs-2310-07521}; or illogical answers, providing incoherent and poorly structured knowledge~\cite{DBLP:journals/corr/abs-2310-00741,DBLP:journals/corr/abs-2302-10198,DBLP:journals/corr/abs-2305-18395}. Although recent method FactTune~\cite{DBLP:journals/corr/abs-2311-08401} improves the factuality of answers by increasing the proportion of correct facts, it ignores other critical aspects, such as completeness~\cite{DBLP:conf/emnlp/MinKLLYKIZH23} and logicality~\cite{DBLP:conf/acl/XuSIC23}.

We hypothesize that these limitations of \acp{LLM} arise from insufficient fine-grained and coarse-grained knowledge awareness during vanilla fine-tuning~\cite{DBLP:journals/corr/abs-2303-16421,DBLP:journals/corr/abs-2310-06271,DBLP:journals/corr/abs-2312-09979,DBLP:journals/corr/abs-2401-17585}. 
On the one hand, as illustrated in Figure~\ref{fig:example}, at the fine-grained level, vanilla fine-tuned \acp{LLM} face difficulties in identifying detailed atomic knowledge within the answer, leading to inadequate awareness of fine-grained knowledge. 
On the other hand, at the coarse-grained level, LLMs frequently fail to distinguish between reliable and unreliable knowledge in answers, indicating a lack of coarse-grained knowledge awareness.
Consequently, there is a pressing need for designing knowledge-aware fine-tuning methods.
This leads to our central research question: \textit{how can we effectively improve both the fine-grained and coarse-grained knowledge awareness of LLMs to address complex knowledge-intensive tasks?}

To this end, we propose a novel knowledge-aware fine-tuning method, named \ac{KnowTuning}, which aims to improve the fine-grained and coarse-grained knowledge awareness of \acp{LLM}. \ac{KnowTuning} consists of two stages: (i) fine-grained knowledge augmentation, and (ii) coarse-grained knowledge comparison. 
In the first stage, we filter difficult atomic knowledge with high perplexity from original answers, and rewrite fine-grained \ac{QA} pairs based on the filtered knowledge. After that, we subsequently use both the original and fine-gained \ac{QA} pairs to train LLMs. 
In the second stage, we adopt several knowledge-disturbing techniques to construct coarse-grained knowledge comparison sets along three dimensions, completeness, factuality, and logicality.
Specifically, we generate answers that are worse in terms of completeness, factuality, or logicality, by deleting, revising, and shuffling the atomic knowledge.
Besides, we rephrase original answers based on the atomic knowledge to prevent overfitting. Finally, we combine the rephrased answers and answers with worse completeness, factuality, and logicality as our knowledge comparison sets. We adopt \acf{DPO}~\cite{DBLP:journals/corr/abs-2305-18290} for optimizing \acp{LLM} on our coarse-grained knowledge comparison sets.

We conduct experiments on a generic \ac{QA} dataset and a medical \ac{QA} dataset using automatic and human evaluations.
Experimental results demonstrate the effectiveness of our proposed method \ac{KnowTuning}, assessing completeness, factuality, and logicality across various sizes of \acp{LLM}.
Furthermore, we demonstrate that KnowTuning not only generates more facts but also reduces the factual error rate during fine-grained facts evaluation.

In summary, our main contributions are:
\begin{itemize}[leftmargin=*,nosep]
    \item We focus on systematically enhancing the knowledge awareness of \acp{LLM} at both fine-grained and coarse-grained levels to address complex knowledge-intensive tasks.
    \item We introduce \ac{KnowTuning}, a novel method that fine-tunes \acp{LLM} to leverage fine-grained knowledge augmentation and coarse-grained knowledge comparison to improve fine-grained and coarse-grained knowledge awareness of \acp{LLM}.
    \item We demonstrate the effectiveness of \ac{KnowTuning} in the generic and medical domain \ac{QA} datasets through automatic and human evaluations, across various sizes of \acp{LLM}. Furthermore, KnowTuning generates more facts with less factual error rate under fine-grained facts evaluation.\footnote{The code is available at \url{https://github.com/youganglyu/KnowTuning}}
\end{itemize}

\begin{figure*}[htbp]
  \centering
\includegraphics[width=1.0\textwidth]{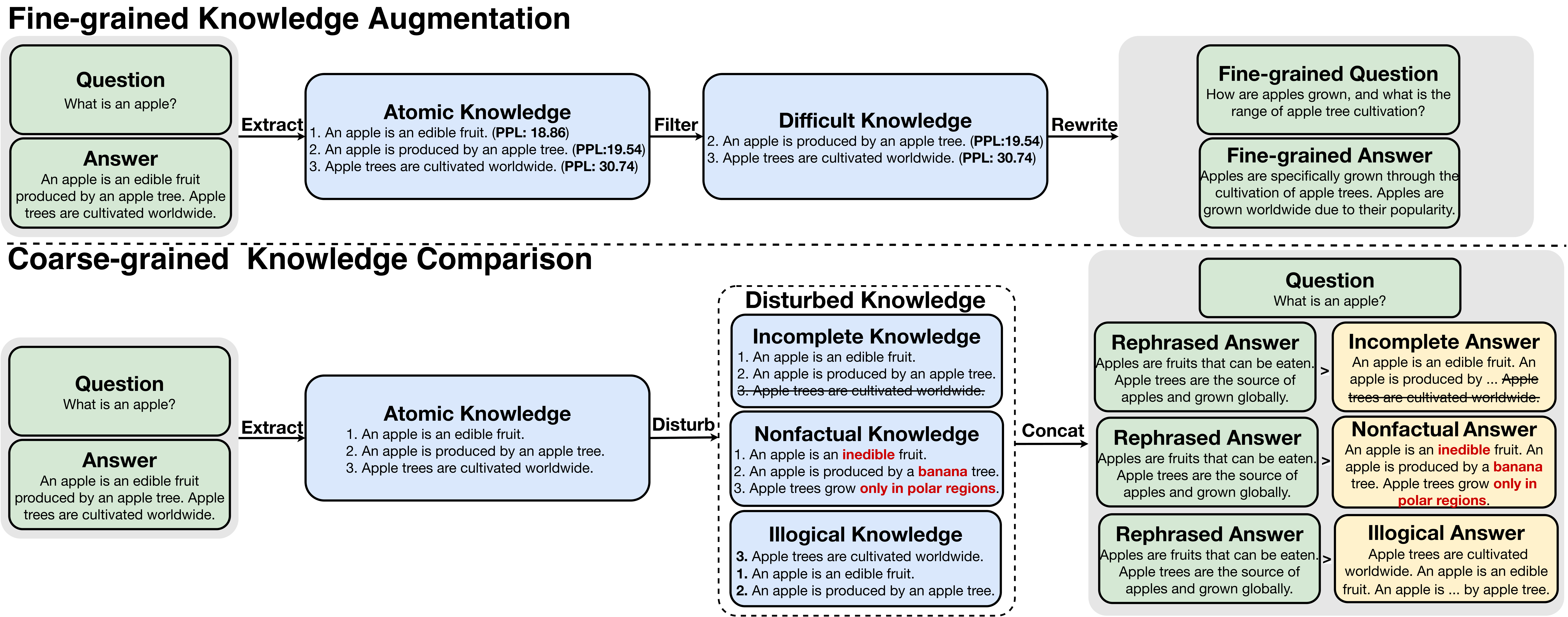}
\vspace*{-1mm}
\caption{Overview of \ac{KnowTuning}. \ac{KnowTuning} leverages fine-grained knowledge augmentation and coarse-grained knowledge comparison to improve the knowledge awareness of \acp{LLM}.} 
\vspace*{-5mm}
\label{fig:model}
\end{figure*}

\section{Related Work}
\label{sec:related_work}

\subsection{\acp{LLM} for Knowledge-intensive Tasks}
\Acfp{LLM} have been applied to various knowledge-intensive tasks~\cite{DBLP:conf/naacl/MoiseevDAJ22,DBLP:conf/iclr/0002IWXJ000023, DBLP:journals/corr/abs-2212-14024,DBLP:journals/corr/abs-2311-08401,DBLP:journals/corr/abs-2311-06503,DBLP:journals/corr/abs-2304-14732,DBLP:journals/corr/abs-2311-05085,DBLP:conf/pkdd/NguyenCNCV23,zhang2024towards}. Previous work mainly focus on knowledge-intensive tasks with short-form answers.
~\citet{DBLP:conf/acl/0010LLWWBCH22} use few-shot demonstrations to elicit relevant knowledge statements from \acp{LLM} for \ac{QA} tasks.~\citet{DBLP:conf/emnlp/0010HLHWHC22} train a neural model to generate relevant knowledge through reinforcement learning for \ac{QA} tasks.~\citet{DBLP:conf/emnlp/0010PH0C23} propose a unified model for generating relevant knowledge and solving \ac{QA} tasks. 

However, these methods primarily address multiple-choice \ac{QA}, rather than the more complex open-ended knowledge-intensive QA tasks~\cite{DBLP:conf/naacl/KrishnaRI21,DBLP:journals/corr/abs-2207-05221,DBLP:conf/emnlp/0010HLHWHC22,DBLP:conf/emnlp/0010PH0C23,DBLP:journals/corr/abs-2305-18395}, which aim to solve questions that require detailed explanations and extensive domain knowledge. Recent research indicates that \acp{LLM} face challenges in tackling complex knowledge-intensive \ac{QA} tasks~\cite{DBLP:journals/corr/abs-2306-09296,DBLP:journals/corr/abs-2310-09725,DBLP:journals/corr/abs-2307-03109}. In particular, they are prone to generating responses that are non-factual~\cite{DBLP:conf/nips/LeePXPFSC22,DBLP:conf/aaai/SunSGRRR23,DBLP:conf/acl/0003LZSJ0F22}, incomplete~\cite{DBLP:journals/corr/abs-2212-13138,DBLP:journals/corr/abs-2303-16421}, or illogical~\cite{DBLP:journals/corr/abs-2310-00741,DBLP:journals/corr/abs-2302-10198}. Recently, for open-ended knowledge-intensive tasks, ~\citet{DBLP:journals/corr/abs-2311-08401} propose a method FacTune to improve factuality. Specifically, they first automatically evaluate the proportion of correct facts in candidate answers as factuality scores, and fine-tuning \acp{LLM} to increase the likelihood of generating answers with higher factuality scores. In contrast, we focus on improving the knowledge awareness of \acp{LLM} at multiple essential aspects simultaneously, for solving complex knowledge-intensive \ac{QA} tasks.

\subsection{Fine-tuning for \acp{LLM}}
Fine-tuning is a kind of method to optimize pre-trained \acp{LLM} for further learning downstream domain knowledge and how to exploit the corresponding knowledge to answer diverse questions~\cite{DBLP:conf/nips/BrownMRSKDNSSAA20,DBLP:conf/nips/Ouyang0JAWMZASR22}. 
Previously, fine-tuning is mainly focused on enhancing general-purpose \ac{QA} abilities of \acp{LLM}~\cite{DBLP:conf/emnlp/WangMAKMNADASPK22,DBLP:conf/iclr/WeiBZGYLDDL22,DBLP:conf/icml/LongpreHVWCTZLZ23}. These approaches mainly adopt human-annotated datasets to build the \ac{QA} dataset. Recently, an alternative strategy involves generating \ac{QA} datasets through the utilization of advanced \acp{LLM} to create answers to a variety of questions~\cite{DBLP:conf/acl/WangKMLSKH23,DBLP:journals/corr/abs-2305-17493}. 

Another line of fine-tuning methods fuse information about the quality of the generated answers into the supervision signals~\cite{DBLP:journals/corr/abs-2305-10425,DBLP:journals/corr/abs-2311-04072,DBLP:journals/corr/abs-2309-02144,DBLP:journals/corr/abs-2304-06767,DBLP:journals/corr/abs-2401-06081,DBLP:conf/coling/ZhaoY0XWMCRY24}. ~\citet{DBLP:journals/corr/abs-2305-18290} propose \acf{DPO} to directly optimize \acp{LLM} on the pair-wise comparison set. \citet{DBLP:journals/corr/abs-2306-17492} propose \ac{PRO}
to fine-tune \acp{LLM} on list-wise comparison sets. ~\citet{DBLP:journals/corr/abs-2304-05302} propose a margin-rank loss to optimize the \acp{LLM} on comparison sets. Since collecting large-scale human judgment for the quality of generated answers is expensive,~\citet{DBLP:journals/corr/abs-2212-08073} and ~\citet{DBLP:journals/corr/abs-2309-00267} propose \ac{RLAIF} methods to leverage off-the-shelf \acp{LLM} to annotate general helpfulness scores. In contrast, our work focuses on enhancing the fine-grained and coarse-grained knowledge-awareness of \acp{LLM} to improve performance in terms of completeness, factuality, and logicality simultaneously.


\section{Method}
\label{sec:method}
In this section, we detail the \ac{KnowTuning} method.  
First, we introduce the preliminaries.
Then, we introduce the fine-grained knowledge augmentation.
Next, we introduce coarse-grained knowledge comparison in detail. Finally, a training process for \ac{KnowTuning} is explained.
\subsection{Preliminaries}
\label{ssec:problem_formulatin}
\header{Supervised fine-tuning} Supervised fine-tuning (SFT) aims to train pre-trained \acp{LLM} to understand and answer natural language questions. Formally, given a \ac{QA} dataset $\mathcal{D}=\{(q_i,a_i)\}_{i=1}^N$, where $q_i$ and $a_i$ denotes a question and a corresponding answer. The training objective of \ac{SFT} is to minimize the following loss: 
\begin{equation} 
\label{eq:1}
\mathcal{L}_\mathrm{sft}=-\sum_{j=1}^{|a_i|}\log P_{\pi_{sft}}(a_{i,j}|a_{i,<j},q_i),
\end{equation}
where $a_{i,j}$ denotes the $j$-th token of $a_{i}$.

\header{Atomic knowledge}
Since individual facts can well cover the knowledge in answers~\cite{DBLP:conf/naacl/NenkovaP04,DBLP:conf/emnlp/ZhangB21,DBLP:conf/acl/LiuFLZNHHJWXR23,DBLP:conf/emnlp/MinKLLYKIZH23,wei2024long}, we break an answer into individual facts as atomic knowledge. The atomic knowledge is a short statement conveying one piece of fact, which is a more fine-grained unit than a sentence. Specifically, we extract atomic knowledge set $\mathcal{K}$ from the original answers $a$ as follows:
\begin{equation} 
\label{eq:2}
\mathcal{K}_{i}=\{k_{i}^{j}\}_{j=1}^{|\mathcal{K}_{i}|}=\operatorname{Extract}(a_{i}),
\end{equation}
where $\operatorname{Extract}(\cdot)$ is implemented by prompting OpenAI models to extract atomic knowledge, following~\citet{DBLP:conf/emnlp/MinKLLYKIZH23}.

\subsection{Fine-grained Knowledge Augmentation}
\label{ssec:triple_knowledge}
As illustrated in Figure~\ref{fig:model}, to improve the fine-grained knowledge awareness of \acp{LLM}, we filter difficult atomic knowledge for \acp{LLM}, and rewrite fine-grained QA pairs based on the difficult knowledge.
After that, we subsequently use both the original and fine-gained QA pairs to train LLMs. 
To filter the difficult atomic knowledge for \acp{LLM}, we first compute the generation perplexity $ppl^{j}_{i}$ of each atomic knowledge $k_{i}^{j}$ conditioned on $q_{i}$ as  follows:
\begin{equation} 
\label{eq:2}
ppl_{i}^{j}=\sqrt[n]{\frac1{\sum_{m=1}^{|k_i^{j}|}P_{\pi_{SFT}}(k_{i,m}^{j}|k_{i,<m}^{j},q_i)}}.
\end{equation}
Since high perplexity $ppl$ indicates the lack of knowledge awareness of \acp{LLM} on specific atomic knowledge, we select $\alpha$ percent of the atomic knowledge set $\mathcal{K}_{i}$ in descending order of perplexity to form the difficult knowledge set 
$\mathcal{K}_{i}^{*}$. 
Then, we rewrite the question $q_i$ as a fine-grained question $q_i^{*}$ relevant to difficult knowledge $\mathcal{K}_{i}^{*}$, as follows:
\begin{equation} 
\label{eq:7}
q^{*}_{i}=\operatorname{Rewrite}(q_{i},\mathcal{K}_{i}^{*}),
\end{equation}
where $\operatorname{Rewrite}(\cdot)$ is implemented by prompting
OpenAI models. In addition, we rewrite the answer based on the difficult knowledge set as the fine-grained answer:
\begin{equation} 
\label{eq:10}
a^{*}_{i}=\operatorname{Rewrite}(\mathcal{K}^{*}_{i}).
\end{equation}
Finally, we combine the original \ac{QA} dataset $\mathcal{D}$ and the fine-grained \ac{QA} pairs as the fine-grained knowledge augmentation dataset $\mathcal{D}_{ka}$ as:
\begin{equation} 
\label{eq:4}
\mathcal{D}_{ka}=\mathcal{D} \cup \{q_{i}^{*},a_{i}^{*}\}_{i=1}^{N}.
\end{equation}

\subsection{Coarse-grained Knowledge Comparison}
\label{ssec:compare_contruct}
To improve coarse-grained knowledge awareness of \acp{LLM} in terms of completeness, factuality and logicality, we construct three comparison sets by deleting, revising, and shuffling atomic knowledge. 

\header{Knowledge completeness comparison} To improve knowledge completeness awareness of \acp{LLM}, we construct the knowledge completeness comparison set by randomly deleting the atomic knowledge. Specifically, we first randomly delete atomic knowledge $k$ in the atomic knowledge set $\mathcal{K}$ as incomplete knowledge set:
\begin{equation} 
\label{eq:5}
\mathcal{K}^{c}_{i}=\operatorname{Delete}(\mathcal{K}_{i}),
\end{equation}
where $\operatorname{Delete}(\cdot)$ refers to randomly delete $\beta$ percent of atomic knowledge $k$. Then, we concatenate leftover atomic knowledge of the incomplete knowledge set as an incomplete answer:
\begin{equation} 
\label{eq:6}
a^{c}_{i}=\operatorname{Concat}(\mathcal{K}^{c}_{i}).
\end{equation}
In addition, to avoid overfitting on the original answers~\cite{DBLP:journals/corr/abs-2310-05914}, we rephrase the original answers based on the original atomic knowledge set as:
\begin{equation} 
\label{eq:7}
a^{r}_{i}=\operatorname{Rewrite}(\mathcal{K}_{i}).
\end{equation}
Finally, we combine the rephrased answer $a^{r}_{i}$ and the incomplete answer $a^{c}_{i}$ into knowledge completeness comparison set as follows:
\begin{equation} 
\label{eq:8}
\mathcal{D}_{kcc}=\{(q_i,(a_i^{r},a_{i}^{c}))\}_{i=1}^N.
\end{equation}

\header{Knowledge factuality comparison} To improve the knowledge factuality awareness of \acp{LLM}, we construct the knowledge factuality comparison set by revising the atomic knowledge as nonfactual atomic knowledge. 
Specifically, we first revise the atomic knowledge set $\mathcal{K}_{i}$ as follows:
\begin{equation} 
\label{eq:9}
\mathcal{K}^{f}_{i}=  \operatorname{Revise}(\mathcal{K}_{i}),
\end{equation}
where $\operatorname{Revise}(\cdot )$ is implemented by prompting
OpenAI models to revise the atomic knowledge to the wrong atomic knowledge. Then, we concatenate all atomic knowledge in the nonfactual knowledge set as:
\begin{equation} 
\label{eq:10}
a^{f}_{i}=\operatorname{Concat}(\mathcal{K}^{f}_{i}).
\end{equation}

\noindent%
Finally, we combine the rephrased answer $a^{r}_{i}$ and the nonfactual answer $a^{f}_{i}$ into knowledge factuality comparison set as follows:
\begin{equation} 
\label{eq:11}
\mathcal{D}_{kfc}=\{(q_i,(a_i^{r},a_{i}^{f}))\}_{i=1}^N.
\end{equation}

\header{Knowledge logicality comparison} To improve the knowledge logicality awareness of \acp{LLM}, we construct the knowledge logicality comparison set by randomly shuffling the atomic knowledge. 
Specifically, we first randomly shuffle all atomic knowledge in the atomic knowledge set $\mathcal{K}$ as the illogical knowledge set:
\begin{equation} 
\label{eq:12}
\mathcal{K}^{l}_{i}=\operatorname{Shuffle}(\mathcal{K}_{i}),
\end{equation}
where $\operatorname{Shuffle}(\cdot)$ is implemented by shuffling the order of all atomic knowledge $k$ in the atomic knowledge set $\mathcal{K}$. Then, we follow the shuffled order to concatenate all atomic knowledge in the illogical knowledge set as an illogical answer:
\begin{equation} 
\label{eq:13}
a^{l}_{i}=\operatorname{Concat}(\mathcal{K}^{l}_{i}).
\end{equation}
Next, we combine the rephrased answer $a^{r}_{i}$ and the illogical answer $a^{l}_{i}$ into knowledge logicality comparison set as follows:
\begin{equation} 
\label{eq:14}
\mathcal{D}_{klc}=\{(q_i,(a_i^{r},a_{i}^{l}))\}_{i=1}^N.
\end{equation}

\noindent%
Finally, we combine the knowledge completeness comparison set, the knowledge factuality comparison set, and the knowledge logicality comparison set as the coarse-grained knowledge comparison set:
\begin{equation} 
\label{eq:15}
\mathcal{D}_{kc}=\mathcal{D}_{kcc} \cup \mathcal{D}_{kfc} \cup \mathcal{D}_{klc}.
\end{equation}
\subsection{Training}
\label{ssec:tuning}
To improve the knowledge awareness of \acp{LLM} for solving complex knowledge-intensive tasks, \ac{KnowTuning} includes fine-grained knowledge augmentation training and coarse-grained knowledge comparison training. Specifically, we first train 
\acp{LLM} on fine-grained knowledge augmentation dataset $\mathcal{D}_{ka}$, resulting in a model denoted as $\pi_{ka}$. To improve the coarse-grained knowledge awareness of the model $\pi_{ka}$, we rewrite the \ac{DPO}~\cite{DBLP:journals/corr/abs-2305-18290} loss as follows:
\begin{eqnarray} 
&\mathcal{L}_{dpo}\! = {}  \!-\mathbb{E}_{(q,(a_w,a_l))\sim\mathcal{D}_{kc}} \bigg [\!\log \sigma \bigg(\!\beta \log\!  \frac{\pi_{kc}(a_w | q)}{\pi_{ka}(a_w | q)} 
\hspace*{-5mm}\mbox{} 
\nonumber \\
 & - \beta \log  \frac{\pi_{kc}(a_l | q)}{\pi_{ka}(a_l | q)} \bigg) \bigg ],
\label{eq:16}
\end{eqnarray}
where $ (a_w, a_l) $ denotes the answer pair of the question $q \in D_{kc}$, and $a_{w}$ is the better answer. To maintain coarse-grained knowledge awareness of better answers, we add SFT loss into the coarse-grained knowledge comparison loss:
\begin{equation} 
\label{eq:17}
\mathcal{L}_{kc}\! = \mathcal{L}_{dpo}+\gamma \mathcal{L}_\mathrm{sft},
\end{equation}
where $\mathcal{L}_\mathrm{sft}$ is a term for better answers $a_{w}$ and $\gamma$ is a scalar weighting hyperparameter.

\section{Experiments}
\label{sec:experimental_setup}
\subsection{Research Questions}
We aim to answer the following research questions in our experiments:
\textbf{RQ1}: How does \ac{KnowTuning} perform on generic and medical \ac{QA} under automatic evaluation and human evaluation?
\textbf{RQ2}: How does \ac{KnowTuning} perform on generic and medical \ac{QA} under fine-grained facts evaluation?
\textbf{RQ3}: How do fine-grained knowledge augmentation and coarse-grained knowledge comparison affect the performance of \ac{KnowTuning}?

\begin{table*}[htbp]
\centering \small
\setlength{\tabcolsep}{4pt}
\begin{tabular}{@{}l cccccccc}
\toprule
&
  \multicolumn{2}{c}{\textbf{Dolly}} &
  \multicolumn{2}{c}{\textbf{MedQuAD}} &
  \multicolumn{2}{c}{\textbf{NQ}} &
  \multicolumn{2}{c}{\textbf{ELI5}}
  \\ \cmidrule(lr){2-3} \cmidrule(lr){4-5} \cmidrule(lr){6-7} \cmidrule(lr){8-9}
    \textbf{Method}     & METEOR   & BERTScore & METEOR   & BERTScore   & METEOR   & BERTScore & METEOR   & BERTScore   \\ \midrule
& \multicolumn{8}{c}{Backbone Language Model: Llama2-7b-base}                                                          \\ \midrule
Base   & 12.29 & 78.07 & 12.79 & 78.44
& \phantom{0}5.10 & 72.70 & \phantom{0}9.09  & 76.05
\\
SFT                           
 & 14.01 & 84.38  & 19.95	& 80.97
& \phantom{0}7.55 & 76.71 & 11.96 & 79.65 \\
RLAIF                           & 17.60 & 85.31 & 20.60	& 83.82
& 10.77 & 79.62 & 13.66 & 80.41 \\ 
FactTune                             & 16.84 & 85.16 & 21.82 & 82.99
& 10.08 & 79.09 & 14.19 & 80.83 \\
\textbf{KnowTuning}                           & \textbf{19.56} & \textbf{86.37} & \textbf{24.71} & \textbf{84.28}
& \textbf{12.22} & \textbf{80.54} & \textbf{16.32} & \textbf{81.74}
\\
 \midrule
& \multicolumn{8}{c}{Backbone Language Model: Llama2-13b-base}                                                         \\ \midrule
Base   & 11.59 & 77.90 & 12.12 & 78.29
& \phantom{0}5.51 & 73.80 & \phantom{0}7.79 & 75.63
\\
SFT                           
 & 15.31 & 84.39 & 19.66 & 82.34
& \phantom{0}8.70 & 78.18 & 12.00 & 81.21 \\
RLAIF                           & 19.03 & 85.43 & 20.37 & 83.13
& 11.79 & 80.30 & 13.61 & 82.06
\\ 
FactTune                             & 18.59 & 85.38 & 21.42 & 83.49
& 11.37 & 80.02 & 13.74 & 82.16
\\
\textbf{KnowTuning}                           & \textbf{20.01} & \textbf{86.32} & \textbf{25.21} & \textbf{84.41}
& \textbf{12.56} & \textbf{80.74} & \textbf{14.45} & \textbf{83.06} \\ \bottomrule
\end{tabular}
\caption{Lexicon-based and semantic-based evaluation on generic and medical \ac{QA}. The best performance is highlighted in \textbf{bold}.}
\vspace*{-1mm}
\label{tab:rq1.12}
\vspace*{-5mm}
\end{table*}

\subsection{Datasets}
We conduct experiments on general domain and domain-specific knowledge-intensive question-answering datasets:
\begin{itemize}[leftmargin=*,nosep]
\item \textbf{Dolly}~\cite{DatabricksBlog2023DollyV2} is a general domain \ac{QA} dataset carefully curated by thousands of human annotators. Since we focus on open-ended generic domain QA, we filter QA pairs of ``open\_qa'' and ``general\_qa'' categories. 
\item \textbf{MedQuAD}~\cite{DBLP:journals/bmcbi/AbachaD19} is a medical domain \ac{QA} dataset, which is collected from 12 National Institutes of Health websites. Following~\citet{DBLP:conf/acl/AugustRS22}, we filter \ac{QA} pairs of the category “Information” for giving detailed information about medical terms.
\end{itemize}
To evaluate the performance across a wider range of knowledge-intensive tasks, we further evaluate generic \ac{QA} models on two representative test sets from \ac{KILT} benchmark~\cite{DBLP:conf/naacl/PetroniPFLYCTJK21}:
\begin{itemize}[leftmargin=*,nosep]
\item \textbf{NQ}~\cite{DBLP:journals/tacl/KwiatkowskiPRCP19} consists of real questions directed to the Google search engine. Every question is paired with a corresponding Wikipedia page that includes a detailed long-form answer and a concise short answer. We filter questions and corresponding long answers as testing \ac{QA} pairs.
\item \textbf{ELI5}~\cite{DBLP:conf/acl/FanJPGWA19} includes a set of question-answer-evidence triples. The questions are complex, and the responses are comprehensive, explanatory, and presented in a free-form style. We filter questions and corresponding answers as testing \ac{QA} pairs.
\end{itemize}
More details of datasets are in Appendix~\ref{appendix:data}.

\subsection{Baselines}
We compare our model with the following baselines:
\begin{itemize}[leftmargin=*,nosep]
\item \textbf{Base} 
denotes that testing  Llama2-base models~\cite{DBLP:journals/corr/abs-2307-09288} under zero-shot setting.
\item \textbf{\ac{SFT}}~\cite{DBLP:conf/nips/Ouyang0JAWMZASR22} represents vanilla fine-tuning backbone \acp{LLM} on \ac{QA} datasets according to Eq.~\ref{eq:1}.
\item \textbf{\ac{RLAIF}} ~\cite{DBLP:journals/corr/abs-2212-08073,DBLP:journals/corr/abs-2309-00267} leverages \acp{LLM} to annotate overall helpfulness scores for candidate answers, and construct overall helpfulness comparison sets based on the scores.

\item \textbf{FactTune} ~\cite{DBLP:journals/corr/abs-2311-08401} constructs factuality comparison sets by calculating the proportion of correct facts in candidate answers. 
\end{itemize}
More details of baselines are in Appendix~\ref{appendix:base}.

\begin{table*}[htbp]
\setlength{\tabcolsep}{5pt}
\centering \small
\begin{tabular}{@{}l @{~~} l cccccccccl@{}}
\toprule
&
&
  \multicolumn{3}{c}{\textbf{Completeness}} &
  \multicolumn{3}{c}{\textbf{Factuality}} &
  \multicolumn{3}{c}{\textbf{Logicality}} &
  \\ \cmidrule(lr){3-5} \cmidrule(lr){6-8}  \cmidrule(lr){9-11}
    \textbf{Method} & \textbf{Dataset}     & Win   & Tie   & Lose  & Win   & Tie   & Lose  & Win   & Tie   & Lose  & Avg. gap  \\ \midrule
&& \multicolumn{10}{c}{Backbone Language Model: Llama2-7b-base}                                                          \\ \midrule
KnowTuning vs Base & \multirow{4}{*}{Dolly}    & \textbf{88.50\rlap{$^{\ast}$}} & \phantom{0}3.00 & \phantom{0}8.50
& \textbf{73.00\rlap{$^{\ast}$}} & 20.00 & \phantom{0}7.00
& \textbf{80.50\rlap{$^{\ast}$}} & 12.00 & \phantom{0}7.50 & \textbf{+73.00}
\\
KnowTuning vs SFT  &                          
 & \textbf{78.50\rlap{$^{\ast}$}} & \phantom{0}5.50 & 16.00
& \textbf{37.00\rlap{$^{\ast}$}} & 46.50 & 16.50
& \textbf{50.50\rlap{$^{\ast}$}} & 34.00 & 15.50 & \textbf{+39.33} \\
KnowTuning vs RLAIF   &                          & \textbf{69.50\rlap{$^{\ast}$}} & \phantom{0}5.00 & 25.50
& \textbf{32.00\rlap{$^{\ast}$}} & 49.00 & 19.00 
& \textbf{46.50\rlap{$^{\ast}$}} & 39.00 & 14.50  & \textbf{+29.67} \\ 
KnowTuning vs FactTune   &                          & \textbf{64.50\rlap{$^{\ast}$}} & 10.00 & 25.50
& \textbf{30.00\rlap{$^{\ast}$}} & 53.00 & 17.00
& \textbf{31.50\rlap{$^{\ast}$}} & 55.50 & 13.00  & \textbf{+23.50} \\
\midrule
KnowTuning vs Base & \multirow{4}{*}{MedQuAD} & \textbf{93.00\rlap{$^{\ast}$}} & \phantom{0}3.00 & \phantom{0}4.00
& \textbf{72.50\rlap{$^{\ast}$}} & 20.50 & \phantom{0}7.00
& \textbf{85.00\rlap{$^{\ast}$}} & \phantom{0}8.50 & \phantom{0}6.50 & \textbf{+77.67}\\
KnowTuning vs SFT &  
& \textbf{81.00\rlap{$^{\ast}$}} & \phantom{0}3.50 & 15.50
& \textbf{46.50\rlap{$^{\ast}$}} & 37.50 & 16.00
& \textbf{64.50\rlap{$^{\ast}$}} & 21.50 & 14.00 & \textbf{+48.83} \\
KnowTuning vs RLAIF   &                         
& \textbf{85.00\rlap{$^{\ast}$}} & \phantom{0}2.50 & 12.50
& \textbf{41.00\rlap{$^{\ast}$}} & 38.50 & 20.50
& \textbf{50.50\rlap{$^{\ast}$}} & 30.00 & 19.50 & \textbf{+41.33} \\
KnowTuning vs FactTune   &                          & \textbf{83.00\rlap{$^{\ast}$}} & \phantom{0}3.50 & 13.50
& \textbf{40.50\rlap{$^{\ast}$}} & 36.50 & 23.00
& \textbf{50.50\rlap{$^{\ast}$}} & 31.50 & 18.00  & \textbf{+39.83} \\
 \midrule
&& \multicolumn{10}{c}{Backbone Language Model: Llama2-13b-base}                                                         \\ \midrule
KnowTuning vs Base & \multirow{4}{*}{Dolly}    & \textbf{85.50\rlap{$^{\ast}$}} & \phantom{0}6.50 & \phantom{0}8.00
& \textbf{66.00\rlap{$^{\ast}$}} & 24.50 & \phantom{0}9.50
& \textbf{81.00\rlap{$^{\ast}$}} & 13.00 & \phantom{0}6.00 & \textbf{+69.67} \\
KnowTuning vs SFT  &                          & \textbf{77.00\rlap{$^{\ast}$}} & \phantom{0}5.00 & 18.00
& \textbf{35.50\rlap{$^{\ast}$}} & 49.50 & 15.00
& \textbf{45.00\rlap{$^{\ast}$}} & 40.00 & 15.00 & \textbf{+36.50} \\
KnowTuning vs RLAIF   &                          & \textbf{73.50\rlap{$^{\ast}$}} & \phantom{0}4.00 & 22.50
& \textbf{33.50\rlap{$^{\ast}$}} & 52.50 & 14.00
& \textbf{46.50\rlap{$^{\ast}$}} & 40.50 & 13.00  & \textbf{+34.67} \\ 
KnowTuning vs FactTune   &                          & \textbf{68.50\rlap{$^{\ast}$}} & \phantom{0}6.50 & 25.00
& \textbf{30.50\rlap{$^{\ast}$}} & 55.00 & 14.50
& \textbf{36.00\rlap{$^{\ast}$}} & 54.00 & 10.00  & \textbf{+28.50} \\
\midrule
KnowTuning vs Base &
  \multirow{4}{*}{MedQuAD} & \textbf{92.50\rlap{$^{\ast}$}} & \phantom{0}2.50 & \phantom{0}5.00
& \textbf{73.50\rlap{$^{\ast}$}} & 17.50 & \phantom{0}9.00
& \textbf{84.00\rlap{$^{\ast}$}} & \phantom{0}8.00 & \phantom{0}8.00 & \textbf{+76.00} \\
KnowTuning vs SFT  & \multicolumn{1}{l}{}      & \textbf{86.50\rlap{$^{\ast}$}} & \phantom{0}3.50 & 10.00
& \textbf{45.50\rlap{$^{\ast}$}} & 41.00 & 13.50
& \textbf{60.00\rlap{$^{\ast}$}} & 31.00 & \phantom{0}9.00 & \textbf{+53.16} \\
KnowTuning vs RLAIF   &                          & \textbf{82.50\rlap{$^{\ast}$}} & \phantom{0}5.00 & 12.50
& \textbf{38.50\rlap{$^{\ast}$}} & 48.00 & 13.50 & \textbf{54.00\rlap{$^{\ast}$}} & 38.50 & \phantom{0}7.50   & \textbf{+47.17} \\ 
KnowTuning vs FactTune   &                          & \textbf{78.00\rlap{$^{\ast}$}} & \phantom{0}4.50 & 17.50 & \textbf{37.00\rlap{$^{\ast}$}} & 47.00 & 16.00
& \textbf{48.50\rlap{$^{\ast}$}} & 39.50 & 12.00  & \textbf{+39.33} \\ \bottomrule
\end{tabular}
\caption{Main results on generic \ac{QA} and medical \ac{QA} datasets evaluated by GPT-4. The scores marked with $\ast$ mean \ac{KnowTuning} outperforms the baseline significantly with $p$-value$< 0.05$ (sign. test), following~\citet{DBLP:conf/acl/GuanMFLDH20}.}
\vspace*{-1mm}
\label{tab:rq1.1}
\vspace*{-5mm}
\end{table*}

\subsection{Evaluation Metrics}
We present our experimental results using two evaluation metrics: automatic evaluation and human-based evaluation. 
Following previous studies~\cite{DBLP:conf/eacl/ClinciuEH21,slobodkin2023don}, we employ two automatic metrics for absolute quality evaluation: the lexicon-based metric METEOR~\cite{banerjee2005meteor} and the semantic-based metric BERTScore~\cite{zhang2019bertscore}. Since recent studies propose that GPT-4 can effectively evaluate the quality of \acp{LLM} answers~\cite{zheng2024judging,DBLP:journals/corr/abs-2305-14387,DBLP:journals/corr/abs-2302-04166}, we also conduct GPT-4 pairwise evaluation. Specifically, given the golden label as a reference, we employ GPT-4 to rate generated answers on three aspects: completeness, factuality, and logicality, on a range of 1 to 10. 
Following~\citet{DBLP:journals/corr/abs-2212-13138,zheng2024judging,DBLP:journals/corr/abs-2312-07398}, we define completeness, factuality and logicality as:
\begin{enumerate*}[label=(\roman*)]
\item \textbf{Completeness}: it examines whether the
answers provide comprehensive and sufficient knowledge to the questions.
\item \textbf{Factuality}: it examines whether the
knowledge in the answers is factually correct.
\item \textbf{Logicality}: it examines whether the
knowledge in the answers is logically structured.
\end{enumerate*} 
Following~\citet{DBLP:journals/corr/abs-2308-12032,DBLP:journals/corr/abs-2307-08701}, we define ``Win-Tie-Lose'' as: 
\begin{enumerate*}[label=(\roman*)]
\item \textbf{Win}: \ac{KnowTuning} wins twice, or wins once and ties once.
\item \textbf{Tie}: \ac{KnowTuning} ties twice, or wins once and loses once.
\item \textbf{Lose}: \ac{KnowTuning} loses twice, or loses once and ties once. 
\end{enumerate*}

We also employ human judgments as the gold standard for assessing the quality of answers. Specifically, human evaluators perform pair-wise comparisons of the top-performing models identified in automatic evaluations. They are presented with a question with a golden answer, and asked to judge two generated answers on three aspects: completeness, factuality, and logicality.

To evaluate the capabilities of LLMs at a fine-grained level, we follow~\citet{DBLP:conf/emnlp/MinKLLYKIZH23} to conduct fine-grained facts evaluation. Specifically, we first break candidate answers into individual facts, and use \textit{gpt-3.5-turbo} to measure the correctness of each fact based on the golden answer as a reference. Following~\citet{DBLP:journals/corr/abs-2311-08401}, we report the number of correct facts ($\#$ Correct), the number of incorrect facts ($\#$ Incorrect), the number of total facts ($\#$ Total) and the proportion of correct facts out of the total number of extracted facts ($\%$ Correct). More details of the evaluation are in Appendix~\ref{appendix:eval}.

\subsection{Implementation Details}
We employ Llama2-base models of different sizes (7b and 13b) as our backbone models for training.
We adopt the Alpaca template~\cite{taori2023stanford} for training and inference.
The OpenAI model used for  $\operatorname{Extract}(\cdot)$, $\operatorname{Rewrite}(\cdot)$ and $\operatorname{Revise}(\cdot)$ is \textit{gpt-3.5-turbo}.
More details of the implementation are in Appendix~\ref{appendix:training}.


\section{Experimental Results and Analysis}
\label{sec:results}
To answer our research questions, we conduct generic domain and medical domain \ac{QA} experiments,  fine-grained facts evaluation, and ablation studies. In addition, we conducted a case study to gain further understanding of the effectiveness of \ac{KnowTuning}.

\begin{table*}[htbp]
\centering \small
\setlength{\tabcolsep}{5pt}
\begin{tabular}{@{}l @{~~} l cccccccccl@{}}
\toprule
 &
   &
  \multicolumn{3}{c}{\textbf{Completeness}} &
  \multicolumn{3}{c}{\textbf{Factuality}} &
  \multicolumn{3}{c}{\textbf{Logicality}} &
  \\ \cmidrule(lr){3-5} \cmidrule(lr){6-8}  \cmidrule(lr){9-11}
    \textbf{Method} & \textbf{Dataset} & Win   & Tie   & Lose  & Win   & Tie   & Lose  & Win   & Tie   & Lose  &  Avg. gap 
    \\ \midrule
&& \multicolumn{10}{c}{Backbone Language Model: Llama2-7b-base}                                                          \\ \midrule
KnowTuning vs FactTune   & Dolly                         & \textbf{61.00\rlap{$^{\ast}$}} & 12.00 & 27.00
& \textbf{28.00\rlap{$^{\ast}$}} & 58.50 & 13.50
& \textbf{33.50\rlap{$^{\ast}$}} & 50.00 & 16.50  & \textbf{+21.83} \\
KnowTuning vs FactTune   &  MedQuAD                        & \textbf{73.00\rlap{$^{\ast}$}} & \phantom{0}9.00 & 18.00
& \textbf{40.00\rlap{$^{\ast}$}} & 43.00 & 17.00
& \textbf{45.50\rlap{$^{\ast}$}} & 36.00 & 18.50 & \textbf{+35.00} \\ \midrule
&&\multicolumn{10}{c}{Backbone Language Model: Llama2-13b-base}                                                         \\ \midrule
KnowTuning vs FactTune   & Dolly                        & \textbf{58.00\rlap{$^{\ast}$}} & 11.00 & 31.00
& \textbf{32.50\rlap{$^{\ast}$}} & 56.50 & 11.00
& \textbf{35.00\rlap{$^{\ast}$}} & 53.00 & 12.00 & \textbf{+23.83} \\
KnowTuning vs FactTune   & MedQuAD     & \textbf{78.00\rlap{$^{\ast}$}} & \phantom{0}6.50 & 15.50
& \textbf{43.00\rlap{$^{\ast}$}} & 45.50 & 11.50
& \textbf{39.00\rlap{$^{\ast}$}} & 45.50 & 15.50 & \textbf{+39.17} \\ \bottomrule
\end{tabular}
\vspace*{-1mm}
\caption{Human evaluation results on generic domain and medical domain \ac{QA} datasets. The scores marked with $\ast$ mean \ac{KnowTuning} surpass FactTune significantly with $p$-value$< 0.05$ (sign. test).}
\vspace*{-2mm}
\label{tab:rq1.2}
\end{table*}

\begin{table*}[htbp]
\centering \small
\setlength{\tabcolsep}{4pt}
\begin{tabular}{@{}l cccccccc}
\toprule
&
  \multicolumn{4}{c}{\textbf{Dolly}} &
  \multicolumn{4}{c}{\textbf{MedQuAD}} 
  \\ \cmidrule(lr){2-5} \cmidrule(lr){6-9}
    \textbf{Method}     & \# Correct $\uparrow$   & \# Incorrect $\downarrow$ & \# Total $\uparrow$  & \% Correct $\uparrow$  & \# Correct $\uparrow$   & \# Incorrect $\downarrow$  & \# Total $\uparrow$ & \% Correct $\uparrow$   \\ \midrule
& \multicolumn{8}{c}{Backbone Language Model: Llama2-7b-base}                                                          \\ \midrule
Base   & \phantom{0}6.15 & 3.62 & \phantom{0}9.77 & 62.94
& \phantom{0}6.54 & 3.42 & \phantom{0}9.96  & 65.66
\\
SFT                           
 & \phantom{0}7.77 & \textbf{1.85} & \phantom{0}9.62 & 80.77
& 16.11 & 1.73 & 17.84 & 90.30 \\
RLAIF                           & 11.23 & 2.10 & 13.33 & 84.25
& 10.86 & 0.95 & 11.81 & 91.96 \\ 
FactTune                             & 11.25 & 1.92 & 13.17 & 85.42
& 12.83 & \textbf{0.83} & 13.66 & 93.92 \\
\textbf{KnowTuning}                           & \textbf{14.40} & 2.36 & \textbf{16.76} & \textbf{85.92}
& \textbf{18.04} & 0.98 & \textbf{19.02} & \textbf{94.85}
\\
 \midrule
& \multicolumn{8}{c}{Backbone Language Model: Llama2-13b-base}                                                         \\ \midrule
Base   & \phantom{0}9.57 & 4.28 & 13.85 & 69.10
& \phantom{0}7.96 & 3.50 & 11.46 & 69.46
\\
SFT                           
 & \phantom{0}9.96 & 2.21 & 12.17 & 81.84
& 16.82 & 1.66 & 18.48 & 91.02 \\
RLAIF                           & 10.72 & 2.16 & 12.88 & 83.23
& 13.01 & 1.16 & 14.17 & 91.81
\\ 
FactTune                             & 12.73 & \textbf{2.12} & 14.85 & 85.72
& 13.02 & \textbf{1.01} & 14.03 & 92.80
\\
\textbf{KnowTuning}                           & \textbf{15.44} & 2.20 & \textbf{17.64} & \textbf{87.53}
& \textbf{19.01} & 1.11 & \textbf{20.12} & \textbf{94.48} \\ \bottomrule
\end{tabular}
\caption{Fine-grained facts evaluation on generic and medical \ac{QA}. The best performance is highlighted in \textbf{bold}.}
\vspace*{-1mm}
\label{tab:rq2}
\vspace*{-5mm}
\end{table*}

\subsection{Main Results (RQ1)}
\label{ssec:multi_judgment_prediction_results}

\header{Automatic evaluation}
Table~\ref{tab:rq1.12} and Table~\ref{tab:rq1.1} present the reference-based GPT-4 evaluation results and absolute quality evaluation results for both generic and medical domain \ac{QA} datasets. Across all metrics, \ac{KnowTuning} outperforms the baseline models in these domains.
Based on the results, we have three main observations: 
\begin{itemize}[leftmargin=*,nosep]
    \item  \textbf{\ac{KnowTuning} demonstrates effectiveness under lexicon-based and semantic-based evaluations.}  As shown in Table~\ref{tab:rq1.12}, our method consistently improves the absolute quality of answers for general and medical QA tasks. Furthermore, these results illustrate the ability of our method to generalize to a wider range of knowledge-intensive datasets, such as NQ and ELI5.

    \item \textbf{\ac{KnowTuning} consistently outperforms baselines in terms of completeness, factuality and logicality, across generic and domain-specific \ac{QA} datasets.} 
    Compared with Base and \ac{SFT}, \ac{KnowTuning} focuses on improving fine-grained and coarse-grained knowledge awareness of \acp{LLM}, which significantly improves the performance. 
    Compared with \ac{RLAIF} and FactTune, \ac{KnowTuning} is more effective in improving the performance of \acp{LLM} on complex knowledge-intensive \ac{QA} in multiple aspects.
    The reason is that \ac{RLAIF} improves the performance by calculating overall helpfulness scores and FactTune focuses on improving the factuality, they ignore improving the knowledge awareness of \acp{LLM} in multiple essential aspects simultaneously. 
    
    \item  \textbf{\ac{KnowTuning} demonstrates effectiveness on \acp{LLM} across different sizes.} We observe that \ac{KnowTuning} consistently improves the performance of \ac{QA} tasks on different scales (7b and 13B) \acp{LLM}. This finding aligns with~\citet{DBLP:journals/corr/abs-2303-16421} and~\citet{mecklenburg2024injecting}: \acp{LLM} learn a lot of generic knowledge during the pre-training stage but still need to learn downstream domain knowledge and explore how to effectively leverage knowledge for solving knowledge-intensive \ac{QA} tasks.
\end{itemize}

\header{Human evaluation}
\label{ssec:multi_judgment_prediction_results}
Human evaluations are crucial for accurately assessing the quality of answers. As shown in Table~\ref{tab:rq1.2}, to facilitate human annotation processes, we focus on comparing \ac{KnowTuning} with the state-of-art baseline FactTune:
\begin{itemize}[leftmargin=*,nosep]
\item Our findings indicate that \ac{KnowTuning} consistently surpasses FactTune in terms of completeness, factuality, and logicality performance across various sizes of \acp{LLM} under human evaluation.
\item \ac{KnowTuning} demonstrates superior performance over \ac{QA} in both generic and medical domain \ac{QA} evaluated by human, in terms of completeness, factuality, and logicality.
\end{itemize}

\begin{table*}[htbp]
\centering \small
\begin{tabular}{@{}lcccccccccl@{}}
\toprule
 &
  \multicolumn{3}{c}{\textbf{Completeness}} &
  \multicolumn{3}{c}{\textbf{Factuality}} &
  \multicolumn{3}{c}{\textbf{Logicality}} &
  \\ \cmidrule(lr){2-4} \cmidrule(lr){5-7} \cmidrule(lr){8-10}
\textbf{Method} & Win   & Tie   & Lose  & Win   & Tie   & Lose  & Win   & Tie   & Lose  & Avg. gap\\ \midrule
-KA\phantom{C} vs KnowTuning    & 32.50 & 20.00 & 47.50
& 16.00 & 57.50 & 26.50
& 12.50 & 61.50 & 26.00  & -13.00                    \\ \midrule
-KCC vs KnowTuning  & 18.50 & 31.00 & 50.50
& 11.00 & 72.50 & 16.50
& 10.50 & 61.50 & 28.00  &  -18.33     
\\
-KFC vs KnowTuning    & 23.00 & 28.50 & 48.50
& \phantom{0}8.50 & 70.50 & 21.00
& 12.00 & 60.50 & 27.50 &  -17.83                    \\
-KLC vs KnowTuning   & 25.50 & 27.50 & 47.00
& 12.00 & 73.00 & 15.00
& \phantom{0}9.50 & 60.00 & 30.50
& -15.17                \\ 
-KC\phantom{C} vs KnowTuning      & 11.50 & \phantom{0}6.00 & 82.50
& 16.00 & 52.00 & 32.00
& 15.50 & 40.50 & 44.00 & -38.50                    \\
\bottomrule
\end{tabular}
\vspace*{-1mm}
\caption{Ablation study evaluated by GPT-4 on the generic \ac{QA} dataset. The backbone model is Llama2-7b-base. -KA indicates the exclusion of fine-grained knowledge augmentation, -KCC indicates the exclusion of completeness comparison, -KFC indicates the exclusion of factuality comparison, -KLC indicates the exclusion of logicality comparison, and -KC indicates the exclusion of all coarse-grained knowledge comparisons.} 
\vspace*{-5mm}
\label{tab:rq3}
\end{table*}

\subsection{Fine-grained Fact Evaluation (RQ2)}
\label{ssec:rq4}
To evaluate the ability of methods to generate correct facts at the fine-grained level, we conduct fine-grained facts evaluation experiments. Based on the results in Table~\ref{tab:rq2}, we have two main observations:
\begin{itemize}[leftmargin=*,nosep]
    \item \textbf{Knowtuning generates answers with a higher proportion of correct facts across various sizes.} Compared to baselines, \ac{KnowTuning} can generate more facts with less factual error rate across different sizes of \acp{LLM}. Although RLAIF and FactTune improve the proportion of correct facts, they ignore fine-grained knowledge augmentation and coarse-grained knowledge completeness awareness. Note that even though FactTune generates fewer incorrect facts, KnowTuning outperforms FactTune on the more critical metric of the percentage of correct facts.
     \item \textbf{KnowTuning generates larger amounts of correct facts across generic and domain-specific \ac{QA} datasets.} Compared to SFT, we observe that KnowTuning consistently generates more correct facts across generic and domain-specific \ac{QA} datasets. However, in the specific medical domain \ac{QA}, RLAIF and FactTune generate fewer correct facts than SFT. 
     This is because LLMs learn a large amount of generic knowledge during the pre-training stage, yet still lack domain-specific knowledge for downstream tasks~\cite{mecklenburg2024injecting}. 
     This underscores the necessity for enhancing fine-grained knowledge awareness in domain-specific, knowledge-intensive QA tasks, as well as the need to improve coarse-grained knowledge awareness across key aspects of completeness, factuality, and logicality.
\end{itemize}

\subsection{Ablation Studies (RQ3)}
\label{ssec:ablation_study}
In Table~\ref{tab:rq3}, we compare \ac{KnowTuning} with several ablative variants. The variants are as follows:
\begin{enumerate*}[label=(\roman*)]
\item \textbf{-KA}: we remove the fine-grained knowledge augmentation. 
\item \textbf{-KCC}: we remove knowledge completeness comparison set.
\item \textbf{-KFC}: we remove knowledge factuality comparison set. 
\item \textbf{-KLC}: we remove knowledge logicality comparison set.
\item \textbf{-KC}: we remove all coarse-grained knowledge comparison sets.
\end{enumerate*}
Our findings are as follows:
\begin{itemize}[leftmargin=*,nosep]
    \item \header{Removing the fine-grained knowledge augmentation} We observe that removing fine-grained knowledge augmentation (-KA) decreases the performance of all three aspects. This indicates that fine-grained knowledge augmentation is effective for improving fine-grained knowledge awareness of \acp{LLM}.
    \item \header{Removing the coarse-grained knowledge comparison} The absence of coarse-grained knowledge comparisons results in substantial performance degradation in knowledge-intensive QA tasks. Specifically, removing the knowledge completeness comparison (-KCC) adversely affects completeness, the elimination of the knowledge factuality comparison (-KFC) undermines factuality, and the removal of the knowledge logicality comparison (-KLC) diminishes logicality. Although deleting and revising atomic knowledge can impact logicality, shuffling has been found more effective in improving coarse-grained logicality for LLMs. Furthermore, removing all coarse-grained knowledge comparison sets (-KC) results in a significant drop in performance across all aspects of the knowledge-intensive QA task.
\end{itemize}

\subsection{Case Study}
\label{ssec:case_study}
We conduct several case studies and find that KnowTuning is more effective at generating complete, factual and logical answers than baselines across various sizes of LLMs. More details of our case study results are in Appendix~\ref{appendix:case_study_detail}.


\section{Conclusions}
\label{sec:conclusion}
In this paper, we focus on improving the knowledge awareness of \acp{LLM} via fine-tuning for complex knowledge-intensive tasks. 
We have proposed \ac{KnowTuning} to fine-tune \acp{LLM} through fine-grained knowledge augmentation and coarse-grained knowledge comparison stages. We have conducted comprehensive experiments on generic and medical domain \ac{QA} datasets, demonstrating the effectiveness of \ac{KnowTuning} through automatic and human evaluations, across various sizes of \acp{LLM}. Moreover, KnowTuning generates more facts with less factual error rate under fine-grained facts evaluation.

\section*{Limitations}
In this study, \ac{KnowTuning} is mainly aimed at generic and medical knowledge-intensive tasks, we plan to adopt KnowTuning to other tasks such as legal domain \ac{QA}~\cite{zhong2020jec,DBLP:journals/ipm/LyuWRRCLLLS22,DBLP:conf/emnlp/LyuH0ZGRCWR23} and mathematical reasoning~\cite{luo2023wizardmath}. Moreover, our efforts have been concentrated on enhancing the knowledge awareness of \acp{LLM} during the fine-tuning stage. Future studies will aim to explore improving knowledge awareness of \acp{LLM} in the pre-training stage~\cite{DBLP:journals/corr/abs-2007-00655}.

\section*{Ethical Considerations}
\ac{KnowTuning} mainly focuses on completeness, factuality, and logicality, but not social bias~\cite{DBLP:journals/sigmod/PitouraTFFPAW17,DBLP:conf/aaai/LyuLYRRZYR23} or the potential for generating harmful or toxic content~\cite{DBLP:journals/corr/abs-2402-09320,hewitt2024model,gao2024towards}. We plan to adopt our method to reduce social bias and harmful content at fine-grained and coarse-grained levels in future work.

\section*{Acknowledgments}
This work was supported by 
the Natural Science Foundation of China (62272274, 62372275, 62102234, 62202271, 62072279), the National Key R\&D Program of China
with grant No.2022YFC3303004, the Natural
Science Foundation of Shandong Province
(ZR2021QF129), the China Scholarship Council under grant number 202306220180, 
the Dutch Research Council (NWO), under project numbers 024.004.022, NWA.1389.20.\-183, and KICH3.LTP.20.006, 
and 
the European Union's Horizon Europe program under grant agreement No 101070212. All content represents the opinion of the authors, which is not necessarily shared or endorsed by their respective employers and/or sponsors.

\bibliography{references}

\appendix
\section*{Appendix}
\section{Details of Datasets}
\label{appendix:data}
\begin{itemize}[leftmargin=*,nosep]
\item \textbf{Dolly}~\cite{DatabricksBlog2023DollyV2}: Given our focus on open-ended generic domain QA, we selected QA pairs specifically categorized under "open\_qa" and "general\_qa" for our dataset. We filter 4,000 QA pairs for training, 200 QA pairs for validation, and 200 QA pairs for testing.

\item \textbf{MedQuAD}~\cite{DBLP:journals/bmcbi/AbachaD19}:
The dataset covers 37 different question types. In this paper, following~\cite{DBLP:conf/acl/AugustRS22}, we filter \ac{QA} pairs of the category ``Information'' for giving definitions and information about medical terms. We filter 4000 \ac{QA} pairs for training, 200 \ac{QA} pairs for validation and 200 \ac{QA} pairs for testing.

\item \textbf{NQ}~\cite{DBLP:journals/tacl/KwiatkowskiPRCP19}: We filter 200 questions and corresponding long answers as testing \ac{QA} pairs from the development set. The length of these long answers ranges from 100 to 500.
\item \textbf{ELI5}~\cite{DBLP:conf/acl/FanJPGWA19}: We filter 200 questions in the test set and the corresponding highest scoring answers as testing \ac{QA} pairs.
\end{itemize}
\section{Details of Baselines}
\label{appendix:base}
\begin{itemize}[leftmargin=*,nosep]
\item \textbf{Base:} We adopt the Alpaca template~\cite{taori2023stanford} for testing the Llama2-base model~\cite{DBLP:journals/corr/abs-2307-09288} under zero-shot setting.
\item \textbf{\ac{SFT}}: We follow standard vanilla fine-tuning loss in Eq.~\ref{eq:1} to train LLMs on original \ac{QA} datasets.
\item \textbf{\ac{RLAIF}}~\cite{DBLP:journals/corr/abs-2212-08073,DBLP:journals/corr/abs-2309-00267}: We leverage \textit{gpt-3.5-turbo} to annotate overall helpfulness scores and construct generic helpfulness comparison sets. We adopt \ac{DPO}~\cite{DBLP:journals/corr/abs-2305-18290} for generic helpfulness comparison sets optimization.

\item \textbf{FactTune}~\cite{DBLP:journals/corr/abs-2311-08401}: We follow~\citet{DBLP:conf/emnlp/MinKLLYKIZH23} to first break each candidate answers into individual facts, and prompt \acp{LLM} to measure the correctness of each fact based on the golden answer as a reference.\footnote{\url{https://github.com/shmsw25/FActScore}\label{factscore}}  Then, we construct factuality comparison sets by the percentage of correct facts. Finally, we adopt \ac{DPO}~\cite{DBLP:journals/corr/abs-2305-18290} for factuality comparison sets optimization.
\end{itemize}

\section{Details of Evaluation}
\label{appendix:eval}
\subsection{GPT-4 Evaluation}
\label{appendix:gpt4}
\begin{figure*}[htbp]
  \centering
\includegraphics[width=0.9\textwidth]{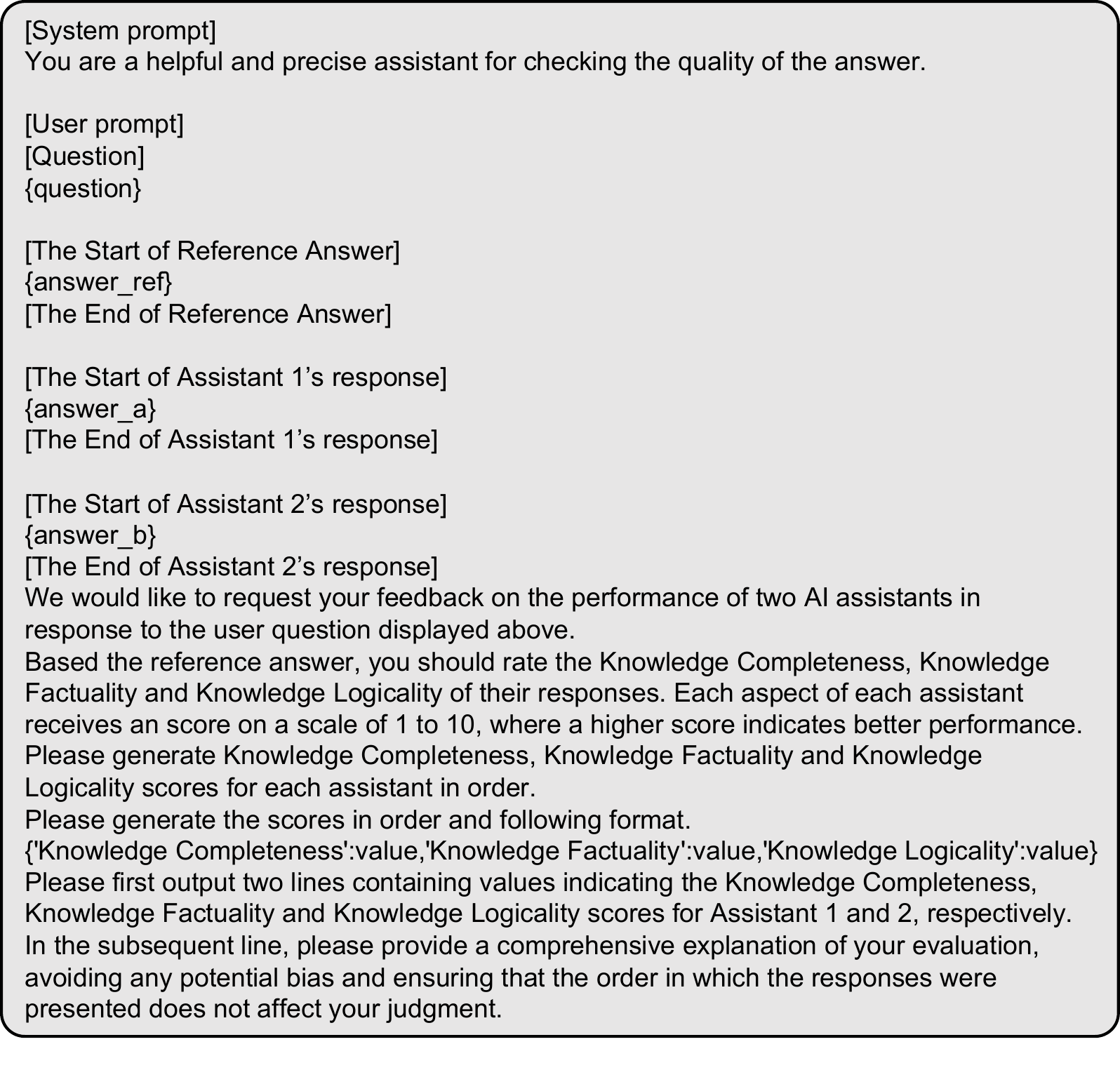}
\vspace*{-1mm}
\caption{Prompts for GPT-4 evaluation.} 
\vspace*{-5mm}
\label{fig:gpt_eval}
\end{figure*}

This section provides specifics of the GPT-4 prompt utilized for reference-based evaluation, employing \textit{gpt4-turbo}. Figure~\ref{fig:gpt_eval} illustrates the adapted prompt from ~\citet{zheng2024judging}, aimed at assessing the completeness, factuality, and logicality of answers. To avoid positional bias~\cite{DBLP:conf/emnlp/KoLKKK20,DBLP:journals/corr/abs-2305-17926}, we evaluate each answer in both positions during two separate runs. 

\begin{figure*}[htbp]
  \centering
\includegraphics[width=0.9\textwidth]{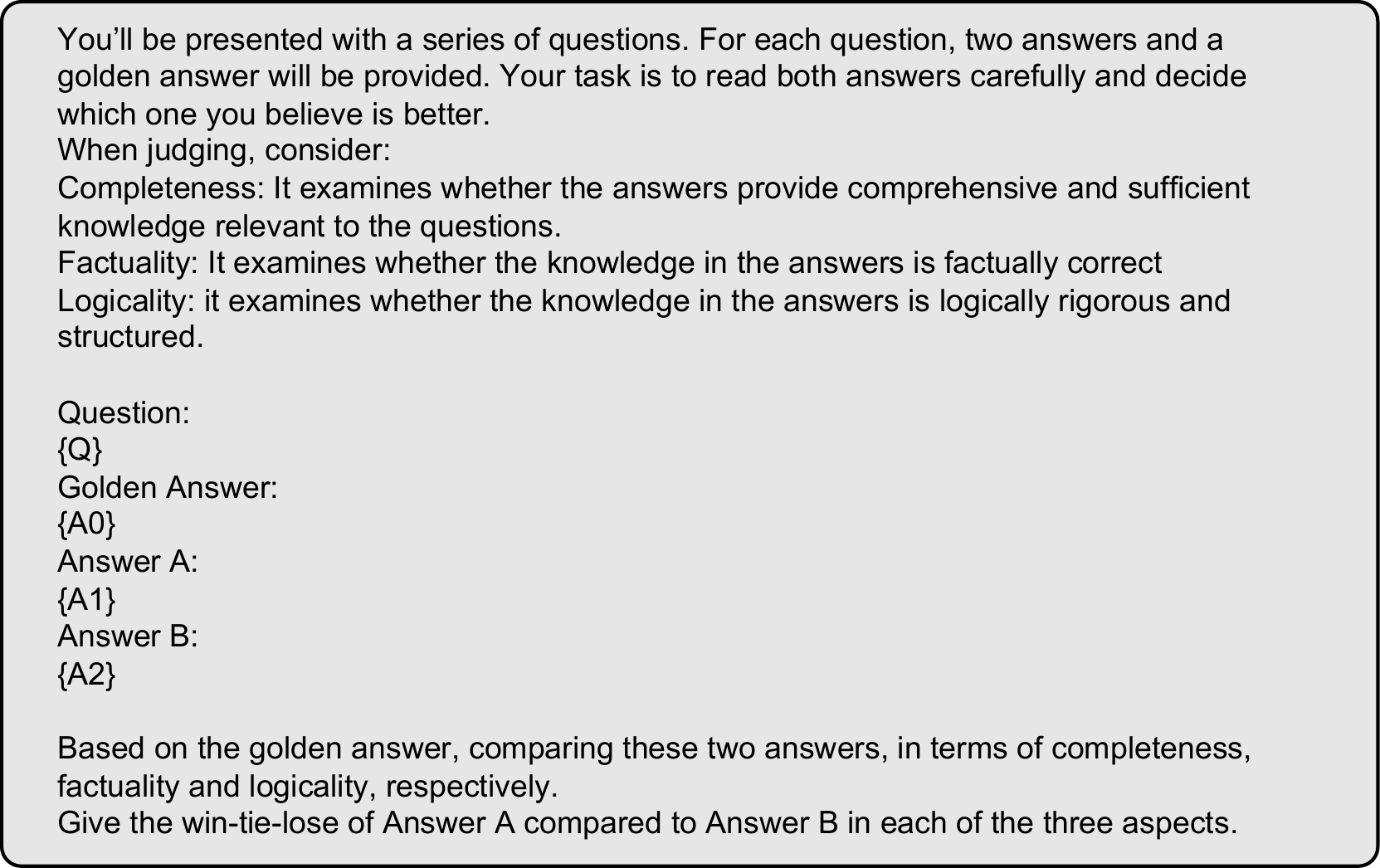}
\vspace*{-1mm}
\caption{Instructions for human evaluation.} 
\vspace*{-5mm}
\label{fig:human_eval}
\end{figure*}

\subsection{Human Evaluation}
\label{appendix:human}
For the human evaluation, we hired people with undergraduate degrees and undergraduate medical degrees to annotate generic QA and medical QA test sets, respectively, to ensure the trustworthiness of the human evaluations, and we allowed the human evaluators to access Wikipedia to further validate the knowledge during the evaluation process. 
Instructions for human evaluation are depicted in Figure~\ref{fig:human_eval}.

\subsection{Fine-grained facts evaluation}
\label{appendix:human}
Following~\citet{DBLP:conf/emnlp/MinKLLYKIZH23}, we first break candidate answers into individual facts, and use \textit{gpt-3.5-turbo} to measure the correctness of each fact based on the golden answer as a reference.\textsuperscript{\ref {factscore}}

\section{Details of Implementation}
\label{appendix:training}
\subsection{Prompts for Extracting, Rewriting, and Revising}
\begin{figure*}[htbp]
  \centering
\includegraphics[width=0.9\textwidth]{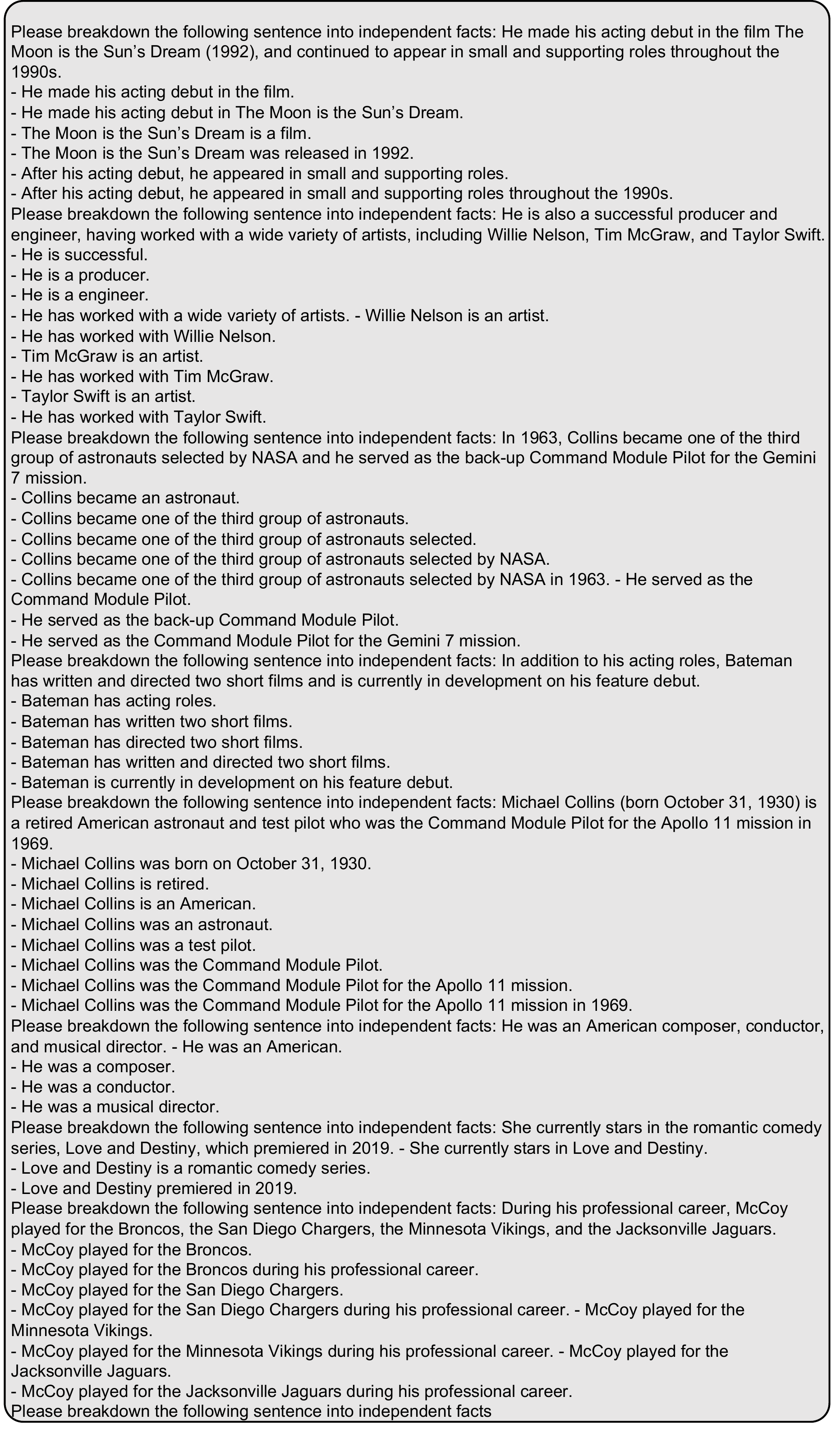}
\vspace*{-1mm}
\caption{Prompts for extracting atomic knowledge in the answer~\cite{DBLP:conf/emnlp/MinKLLYKIZH23}.} 
\vspace*{-5mm}
\label{fig:gpt_extract}
\end{figure*}

\begin{figure*}[htbp]
  \centering
\includegraphics[width=0.9\textwidth]{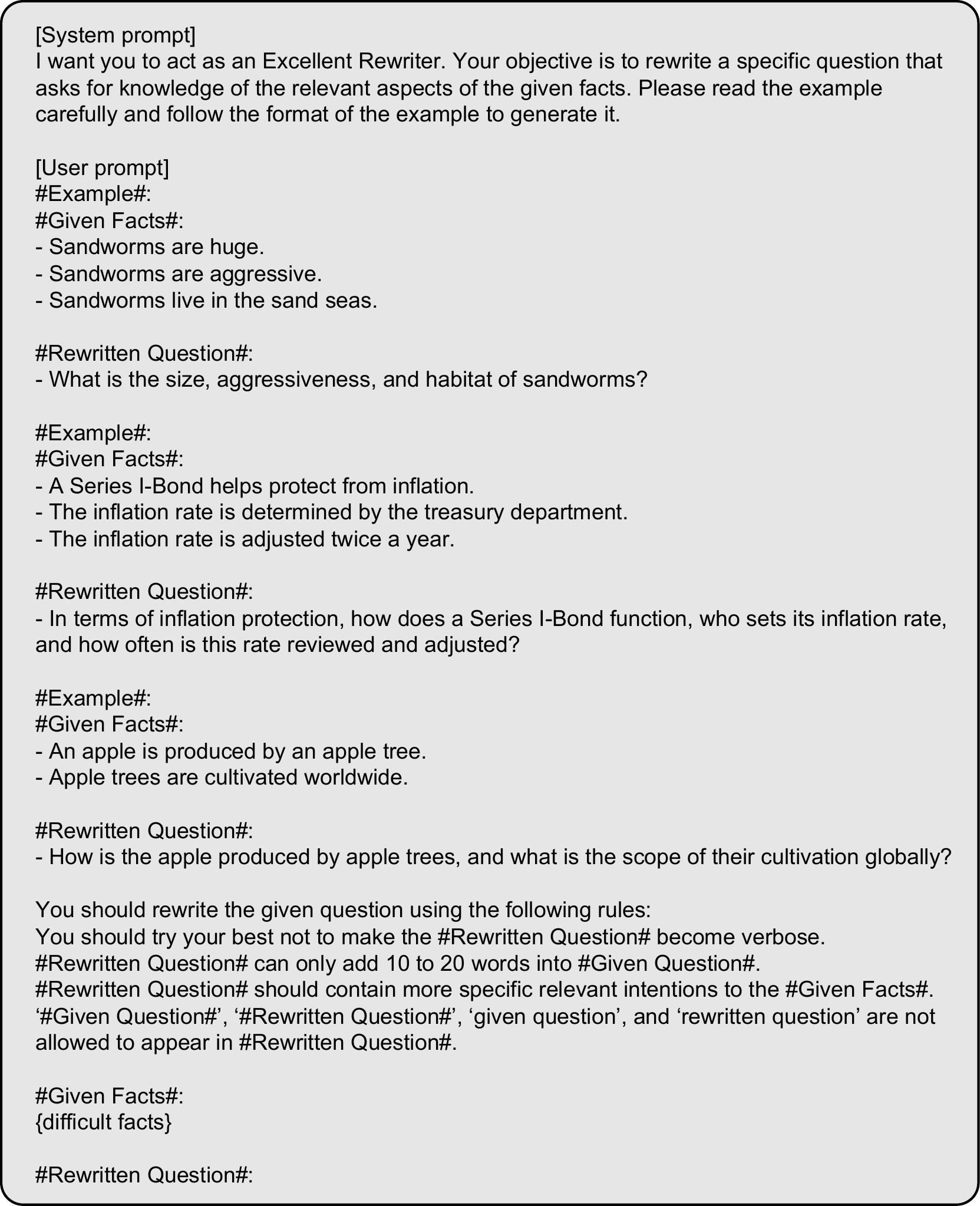}
\vspace*{-1mm}
\caption{Prompts for rewriting fine-grained questions.} 
\vspace*{-5mm}
\label{fig:gpt_rewrite_q}
\end{figure*}

\begin{figure*}[htbp]
  \centering
\includegraphics[width=0.9\textwidth]{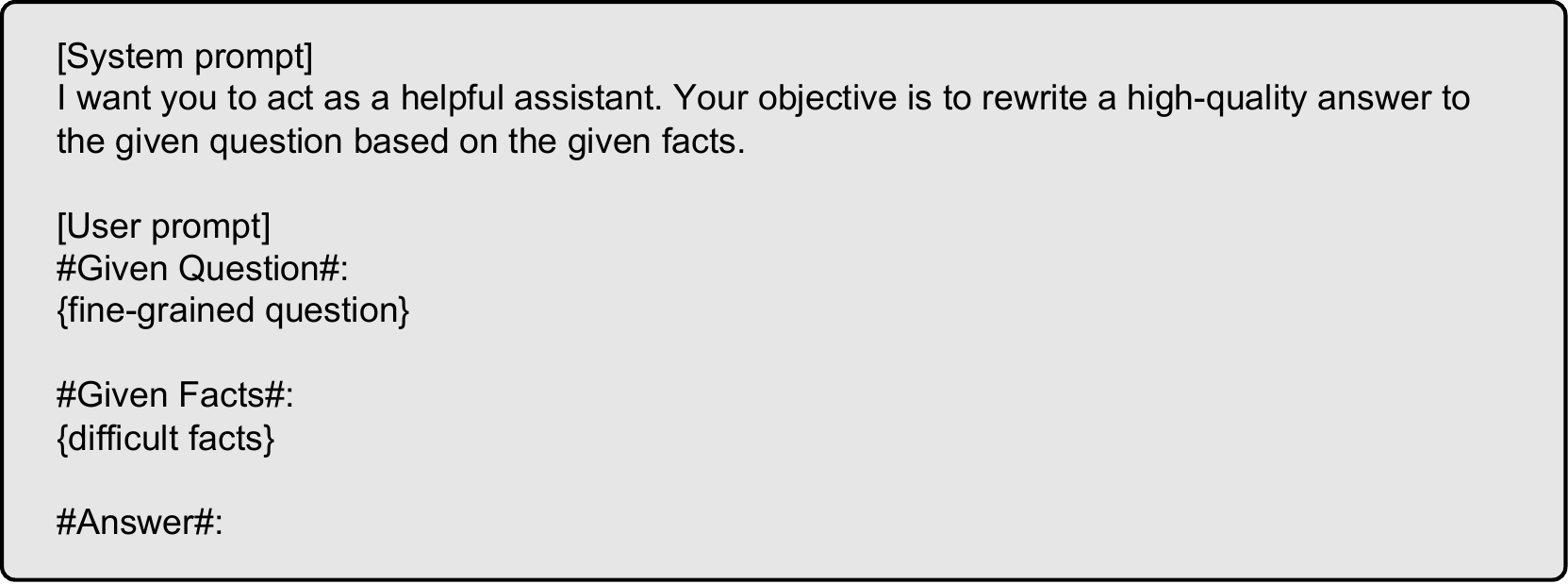}
\vspace*{-1mm}
\caption{Prompts for rewriting fine-grained answers.} 
\vspace*{-5mm}
\label{fig:gpt_rewrite_a}
\end{figure*}

\begin{figure*}[htbp]
  \centering
\includegraphics[width=0.9\textwidth]{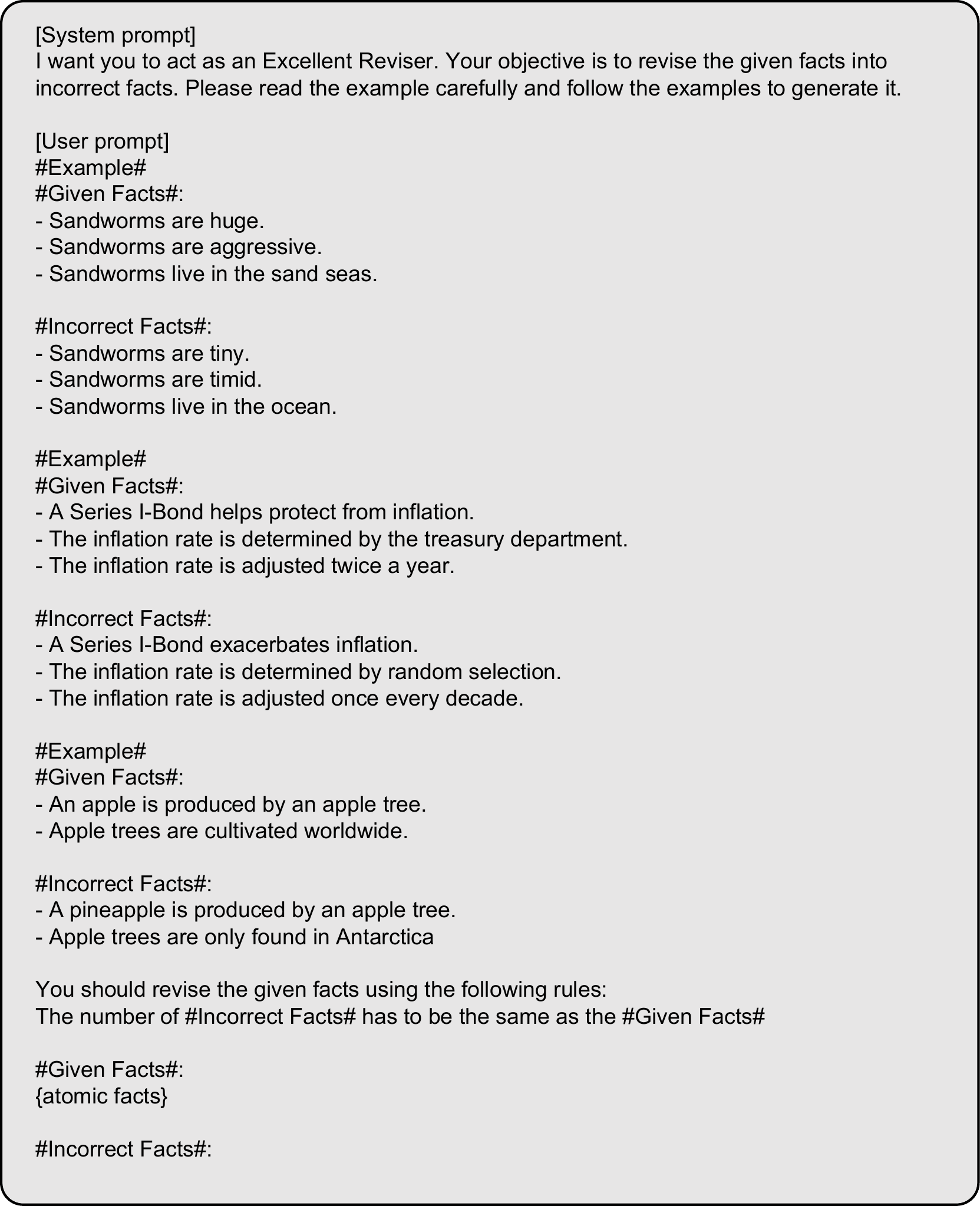}
\vspace*{-1mm}
\caption{Prompts for revising atomic facts into incorrect facts.} 
\vspace*{-5mm}
\label{fig:gpt_revise}
\end{figure*}

Details for the prompts used in $\operatorname{Extract}(\cdot)$, $\operatorname{Rewrite}(\cdot)$, and $\operatorname{Revise}(\cdot)$ are provided. Figures~\ref{fig:gpt_extract}, \ref{fig:gpt_rewrite_q}, \ref{fig:gpt_rewrite_a} and~\ref{fig:gpt_revise} display the prompts for extracting atomic knowledge, rewriting fine-grained questions, rewriting fine-grained answers, and revising atomic knowledge into nonfactual knowledge, respectively.

\subsection{Reliability of atomic knowledge extraction}
To evaluate the reliability of atomic knowledge extraction, we first sample 50 instances of generic\ac{QA} dataset Dolly. We manually checked these data and find that only 3 instances required further separation or merging of atomic facts, illustrating the reliability of extracting atomic facts using \textit{gpt3.5-turbo}.

\subsection{Training}
During the training phase, the AdamW optimizer \cite{DBLP:conf/iclr/LoshchilovH19} is utilized with initial learning rates of $5\cdot 10^{-5}$ for \ac{SFT} and $1\cdot 10^{-5}$ for \ac{DPO}. The batch sizes for \ac{SFT} and \ac{DPO} are set to 32 and 16, respectively, with \ac{SFT} undergoing 3 epochs of training and \ac{DPO} 1 epoch. The filtering and deleting percentages, $\alpha$ and $\beta$, are both fixed at 0.5. The scalar weighting hyperparameter $\gamma $ is set to 0.2. We determine the hyperparameters through pilot experiments. Training leverages PEFT~\cite{peft}, LLaMA-Factory~\cite{llama-factory} and LoRA~\cite{DBLP:conf/iclr/HuSWALWWC22}.

\subsection{Cost Analysis}
The cost of KnowTuning is lower than that of the baseline methods RLAIF and FactTune. Specifically, in the generic domain QA dataset Dolly, the costs are as follows: KnowTuning is \$8.45, RLAIF is \$9.94, and FactTune is \$10.53. This cost difference arises because RLAIF necessitates pairwise comparisons for assessing the overall helpfulness of all candidate answers, while FactTune requires a detailed factuality evaluation for each fact across all candidate answers, thereby increasing their dataset comparison construction costs.

\begin{figure*}[htbp]
  \centering
\includegraphics[width=0.8\textwidth]{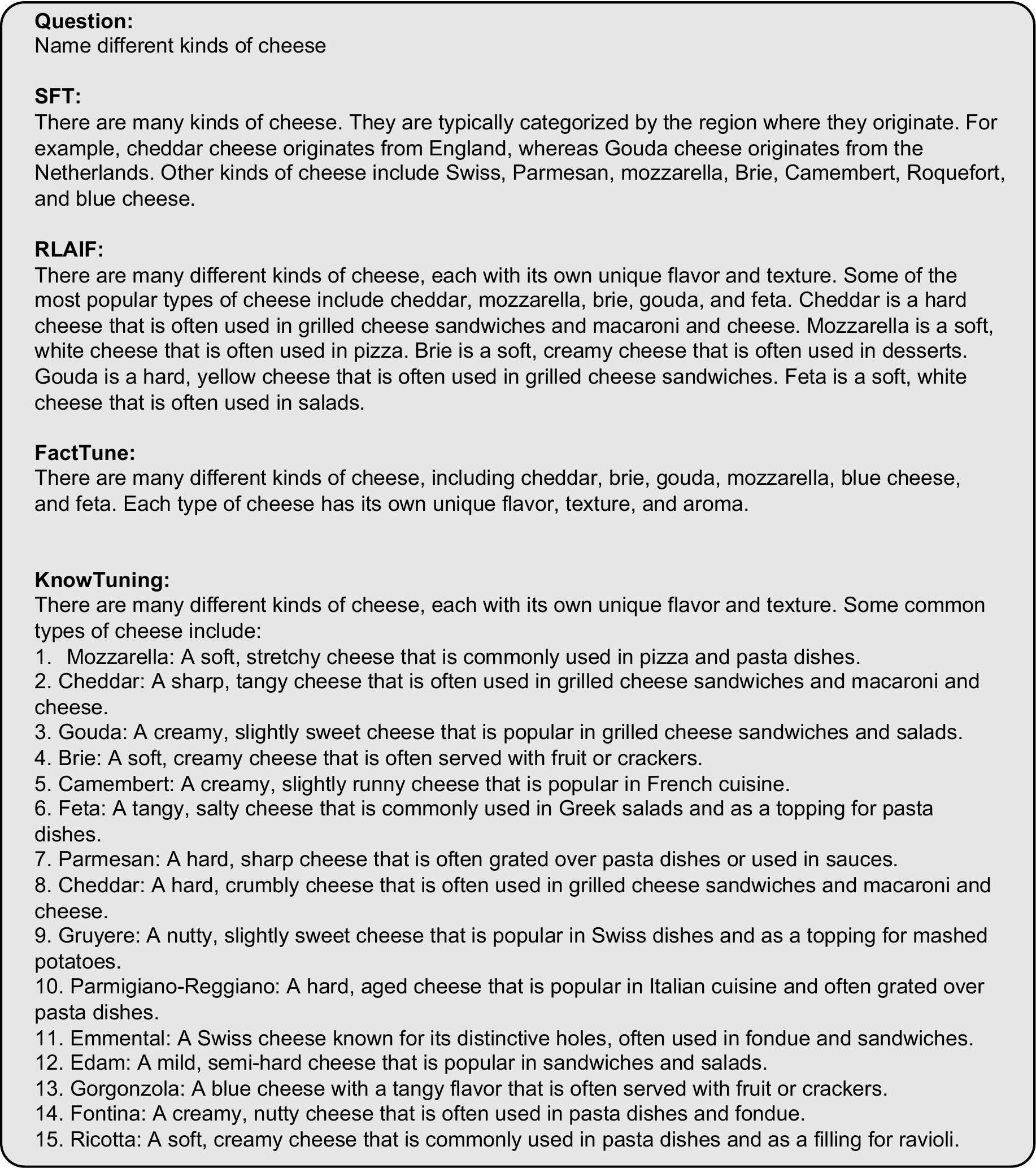}
\vspace*{-1mm}
\caption{Case study for intuitive comparisons on generic \ac{QA} dataset based on Llama2-7b-base.} 
\vspace*{-5mm}
\label{fig:case_study1}
\end{figure*}

\begin{figure*}[htbp]
  \centering
\includegraphics[width=0.8\textwidth]{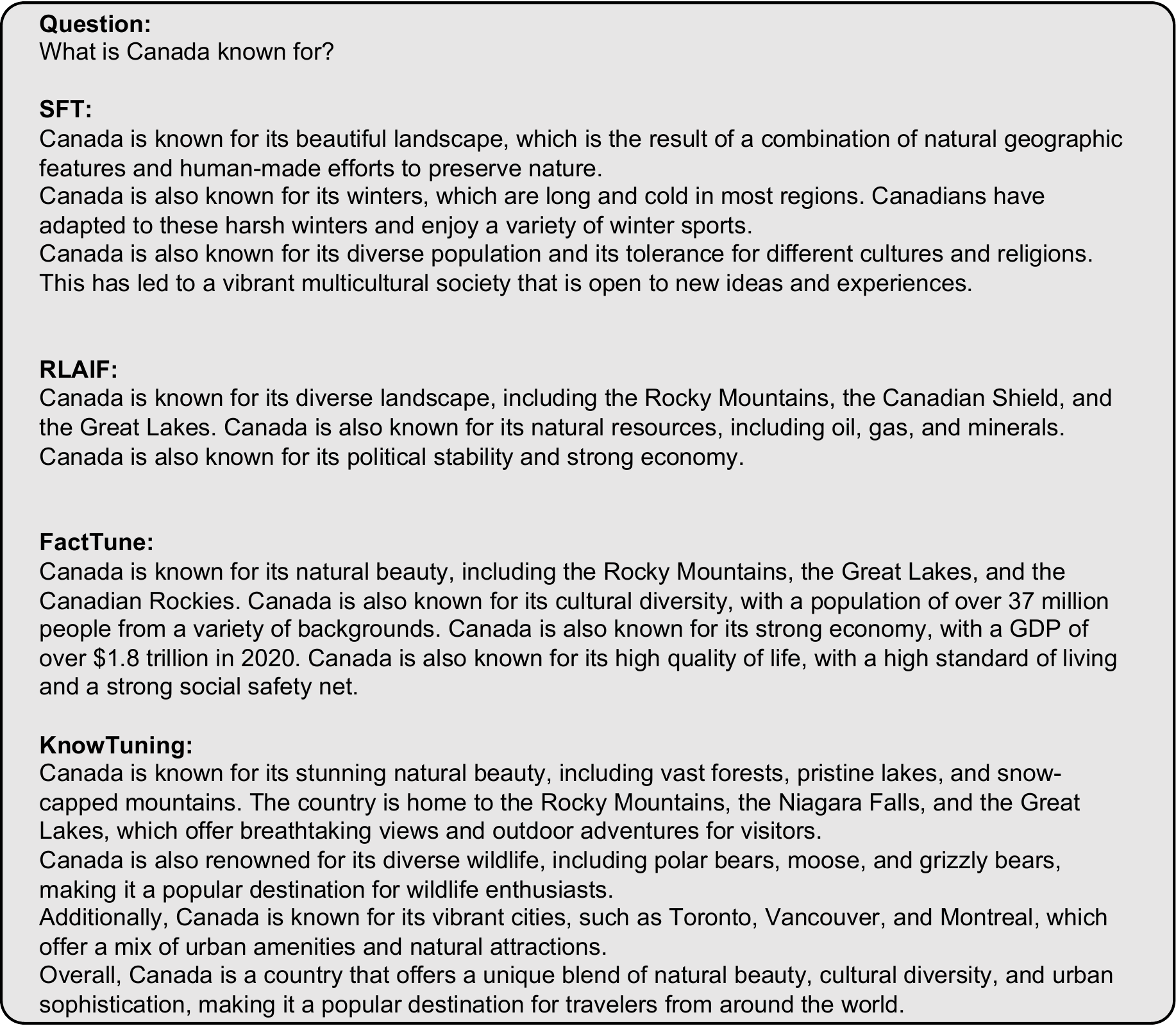}
\vspace*{-1mm}
\caption{Case study for intuitive comparisons on generic \ac{QA} dataset based on Llama2-13b-base.} 
\vspace*{-5mm}
\label{fig:case_study2}
\end{figure*}

\section{Details of Case Study}
\label{appendix:case_study_detail}
As illustrated in Figures~\ref{fig:case_study1} and \ref{fig:case_study2}, the case studies evaluate answers generated by four methods: SFT, RLAIF, FactTune, and KnowTuning across various sizes. Our findings indicate that KnowTuning excels at producing answers that are more complete, factual, and logical across various sizes of LLMs, as detailed below:
\begin{itemize}[leftmargin=*,nosep]
\item As shown in Figure~\ref{fig:case_study1} for the case study based on backbone Llama2-7b-base, KnowTuning generates more complete and logical answers compared to all baselines. Although RLAIF produces more knowledge compared to SFT, it results in fewer logical answers because it does not explicitly focus on logicality optimization. FactTune, on the other hand, focuses on improving the percentage of factualness and performs poorly in terms of answer completeness and logic. This illustrates the need for multiple aspects of coarse-grained knowledge awareness.
\item As shown in Figure~\ref{fig:case_study2} for the case study based on backbone Llama2-13b-base, KnowTuning generates content that is more informative and factual, and the logic between the knowledge is more logical. Although RLAIF generates multiple aspects of knowledge, it does not provide fine-grained knowledge in the answer. FactTune generates detailed information such as Canada's domestic population and GDP, but it provides factually incorrect information. This further underscores the critical need for enhanced fine-grained knowledge awareness.
\end{itemize}
\end{document}